\RequirePackage{fix-cm}
\documentclass[twocolumn]{svjour3}          % twocolumn
\smartqed  % flush right qed marks, e.g. at end of proof
\usepackage{mathptmx}      % use Times fonts if available on your TeX system
\DeclareMathAlphabet{\mathcal}{OMS}{cmsy}{m}{n} % to change the \mathcal back to default after using mathptmx
\usepackage{latexsym}
\usepackage{graphicx}
\usepackage{pbox}
\usepackage{amsfonts,amssymb}
\usepackage{amsmath}
\usepackage{changepage}
\usepackage{tikz}
\usepackage[round,sort]{natbib}
\bibpunct{(}{)}{;}{a}{}{;}
\usepackage{soul}
\usepackage{multirow}
\usepackage{booktabs}
\usepackage{threeparttable}
\usepackage{floatrow}
\usepackage{algorithm}
\usepackage{algcompatible}
\usepackage{tabularx}
\usepackage[titletoc,title]{appendix}
\usepackage{hyperref}
\hypersetup{colorlinks,linkcolor={blue},citecolor={blue},urlcolor={blue}}
\usepackage{url}
\usepackage{bm}

\usepackage[ampersand]{easylist}
\def\setbibindent{
\setlength{\itemindent}{-\bibhang}
\setlength{\leftmargin}{\bibhang}}
\let\urbibitem=\bibitem
\def\bibitem{\setbibindent\urbibitem}
\newcommand{\tikzcircle}[2][black,fill=black]{\tikz[baseline=-0.5ex]\draw[#1,radius=#2] (0,0) circle ;}%

\begin{document}

\title{Zero-Shot Visual Recognition via Bidirectional Latent Embedding
%\thanks{Grants or other notes
%about the article that should go on the front page should be
%placed here. General acknowledgments should be placed at the end of the article.}
}
%\subtitle{Do you have a subtitle?\\ If so, write it here}

%\titlerunning{Short form of title}        % if too long for running head

\author{Qian Wang         \and
        Ke Chen  %etc.
        }

%\authorrunning{Short form of author list} % if too long for running head

\institute{Qian Wang and
           Ke Chen (corresponding author) \at
              School of Computer Science \\
              The University of Manchester\\
               Manchester M13 9PL, UK \\
              \email{\{Qian.Wang, Ke.Chen\}@manchester.ac.uk}           %  \\
%             \emph{Present address:} of F. Author  %  if needed
           %\and
%           Ke Chen (corresponding author) \at
%              School of Computer Science \\
%              The University of Manchester\\
%               Manchester M13 9PL, UK\\
%              \email{Ke.Chen@manchester.ac.uk}
}

\date{Received: date / Accepted: date}
% The correct dates will be entered by the editor

\maketitle

\begin{abstract}
Zero-shot learning for visual recognition, e.g., object and action recognition, has recently attracted a lot of attention. However, it still remains challenging in bridging the semantic gap between visual features and their underlying semantics and transferring knowledge to semantic categories unseen during learning. Unlike most of the existing zero-shot visual recognition methods, we propose a stagewise bidirectional latent embedding framework to two subsequent learning stages for zero-shot visual recognition. In the bottom-up stage, a latent embedding space is first created by exploring the topological and labeling information underlying training data of known classes via  a proper supervised subspace learning algorithm and the latent embedding of training data are used to form landmarks that guide embedding semantics underlying unseen classes into this learned latent space. In the top-down stage, semantic representations of unseen-class labels in a given label vocabulary are then embedded to the same latent space to preserve the semantic relatedness between all different classes via our proposed semi-supervised Sammon mapping with the guidance of landmarks. Thus, the resultant latent embedding space allows for predicting the label of a test instance with a simple nearest-neighbor rule. To evaluate the effectiveness of the proposed framework, we have conducted extensive experiments on four benchmark datasets in object and action recognition, i.e., AwA, CUB-200-2011, UCF101 and HMDB51. The experimental results under comparative studies demonstrate that our proposed approach yields the state-of-the-art performance under inductive and transductive settings.

\keywords{Zero-shot learning \and Object recognition \and Human action recognition \and Supervised locality preserving projection \and Landmark-based Sammon mapping
\and multiple visual and semantic representations}
% \PACS{PACS code1 \and PACS code2 \and more}
% \subclass{MSC code1 \and MSC code2 \and more}
\end{abstract}

\section{Introduction}
\label{intro}
% the background of object and human action recognition as well as the necessity of zero-shot learning.
Visual recognition refers to various tasks for understanding the content of images or video clips. \emph{Object recognition} and \emph{human action recognition} are two typical visual recognition tasks studied extensively in computer vision community. In the last decade, substantial progresses have been made in object and human action recognition \citep{andreopoulos201350}. As a result, we witness a boost of various benchmarks released with more and more classes, which poses greater challenges to computer vision. For example, the number of classes in object recognition benchmarks has increased from 256 in Caltech-256 \citep{griffin2007caltech} to 1000 in ImageNet ILSVRC \citep{russakovsky2015imagenet}, while the number of classes in human action recognition has increased from 51 in HMDB51 \citep{kuehne2011hmdb} to 101 in UCF101 \citep{soomro2012ucf101}. Despite the increasing number of classes in consideration, they are still a small portion of all classes existing in real world. According to \citep{lampert2014attribute}, humans can distinguish approximately 30,000 basic object classes, and much more subordinate ones. Nowadays, new objects emerge rapidly. Practically, it is impossible to collect and annotate visual data for all the classes to establish a visual recognition system.
%On the other hand, the conventional visual recognition methods are based on supervised learning that has a high demand for labelled instances in each class for training.
This leads to a great challenge for visual recognition.

\par
%The introduction of zero-shot learning.
To fight off this challenge, \emph{zero-shot learning} (ZSL) was recently proposed and applied in both object and human action recognition with promising performances, e.g., \citep{akata2013label, akata2014zero, akata2015evaluation, lampert2014attribute, xu2015zero, fu2015transductive, akata2016label, zhang2016zero, zhang2016eccv, changpinyo2016synthesized, zhang2015zero, xian2016latent, al2015transfer, gan2016learning, kodirov2015unsupervised, mensink2014costa, norouzi2013zero,  changpinyo2016predicting, romera2015embarrassingly}.
Unlike the traditional methods that can only recognize classes appearing in the training data, ZSL is inspired by the learning mechanism of human brain and aims to recognize new classes unseen during learning by exploiting intrinsic semantic relatedness between known and unseen classes.
%For example, one can recognize a new species of animal after being told what it looks like and how it is similar to or different from other known animals. It implies that humans can explore the relatedness between different objects from side information (e.g., a high-level description of objects or human actions), and transfer the knowledge from known classes to new ones. Likewise, ZSL exploits intrinsic semantic relatedness between known and unseen classes.
In general, three fundamental elements are required in ZSL; i.e., \emph{visual representation} conveying non-trivial yet informative visual features, \emph{semantic representation} reflecting the relatedness between different classes (especially between known and unseen classes), and  \emph{learning model} properly relating visual features to underlying semantics.

% The visual representations
Visual representations play an important role in visual recognition. In particular, the visual representations learned with deep Convolutional Neural Networks (CNNs) have improved the performances of object recognition, e.g., \citep{chatfield2014return, simonyan2014very, szegedy2015going, he2015deep}, and human action recognition, e.g., \citep{simonyan2014two, zhao2015pooling, wu2015fusing, wang2015towards}. Benefitting from deep learning, zero-shot visual recognition performances have also been boosted, e.g.,  \citep{reed2016learning, al2015transfer, akata2014zero}. In addition, it has been reported that the joint use of multiple visual representations can improve the performances and the robustness of visual recognition, e.g., \citep{fu2015transductive, shao2016kernelized}.
%In this paper, we will use the visual representations which have been widely used in existing ZSL literature.

% semantic representations
Semantic representations aim to model the semantic relatedness between different classes. A variety of semantics modelling techniques  \citep{lampert2014attribute, THUMOS14, liu2011recognizing, mikolov2013distributed, frome2013devise, elhoseiny2015tell, mensink2014costa} have been developed, e.g., semantic attributes \citep{lampert2014attribute, THUMOS14, liu2011recognizing} and word vectors \citep{mikolov2013distributed, frome2013devise}. Semantic attributes are usually manually defined for semantic labels that describe objects and actions contained in images and video streams, while word vectors are automatically learned from unstructured textual data in an unsupervised way. %\citep{mikolov2013distributed}.
%In addition, other side information may also be used to model the relatedness among different classes for ZSL \citep{elhoseiny2015tell, mensink2014costa}.

%The current models for ZSL and their weakness
Given the low-level visual representations of images or video streams and their underlying high-level semantics, the central problem in zero-shot visual recognition is how to transfer knowledge from the visual data of known classes to those of unseen classes. A variety of zero-shot visual recognition methods have been proposed, e.g., \citep{akata2013label, akata2014zero, akata2015evaluation, lampert2014attribute, xu2015zero, fu2015transductive, akata2016label, zhang2015zero, zhang2016zero, zhang2016eccv, changpinyo2016synthesized,  xian2016latent, al2015transfer, gan2016learning, kodirov2015unsupervised, mensink2014costa, norouzi2013zero,  changpinyo2016predicting, romera2015embarrassingly}. A brief review on zero-shot visual recognition will be described in the next section.

%weak points of existing methods
%While a lot of efforts have been made for ZSL including zero-shot visual recognition, the \emph{semantic gap} is still the biggest hurdle hindering the ZSL progress.
In zero-shot visual recognition, the \emph{semantic gap} is the biggest hurdle; i.e., the distribution of instances in visual space is often distinct from that of their underlying semantics in semantic space as visual features in various forms may convey the same concept.
%That is, only visual information (e.g., colour, appearances, motions, etc.) is retained in visual space, while underlying semantics of a class label and the semantic relatedness between different classes have to be described with other information sources such as properties of objects, animal habitats, human culture and so on.
This semantic gap results in a great difficulty in transferring knowledge on known classes to unseen classes.
%Thus, learning a direct mapping or a common space does not address this issue well since the use of supervised learning on labelled instances in known classes does not sufficiently explore intrinsic structures of visual data and the resultant models hence yield poor generalization to test instances in unseen classes.
Apart from the semantic gap issue, the \emph{hubness} \citep{radovanovic2010hubs} is recently identified as a cause that accounts for the poor performance of most existing ZSL models \citep{dinu2014improving, shigeto2015ridge, xu2015zero}. ``Hubness" refers to the phenomenon that some instances (referred to as \textit{hubs}) in the high-dimensional space appear to be the nearest neighbors of a large number of instances. When nearest-neighbour based algorithms are applied, test instances are likely to be close to those ``hubs" regardless of their labels and hence incorrectly labeled as labels of ``hubs". In ZSL, the ``hubness" phenomenon becomes more severe. Apart from the intrinsic property of high-dimensional space \citep{radovanovic2010hubs}, the hubness is exacerbated by a lack of training instances belonging to unseen classes in visual domain and the \emph{domain shift} problem, where the distribution of training data is different from that of test data, which often occurs in ZSL \citep{fu2015transductive, zhang2016eccv}.

%Our solution
In this paper, we propose a novel zero-shot visual recognition framework towards bridging the semantic gap and tackling the hubness issue. Unlike most of existing methods, our framework consists of two subsequent stages: bottom-up and top-down stages. In the bottom-up stage, a latent space is learned from a visual representation
%(or a kernel representation space in the presence of multiple representations)
via supervised subspace learning that preserves intrinsic structures of visual data and promotes the discriminative capability.
%e.g., supervised locality preserving projection (SLPP) \citep{cheng2005supervised} used in our experiments.
We expect that the latent space resulting from such subspace learning captures the intrinsic structures underlying visual data and narrows the semantic gap between visual and semantic spaces.
%so that a learned projection can be extended to test instances in unseen classes.
After the bottom-up learning, in the latent space, the mean of projected points of training data in the same class forms a \emph{landmark} specified as the embedding point of the corresponding class label. In the top-down stage, the semantic representations of all unseen-class labels in a given vocabulary are then embedded in the same latent space (created in the bottom-up stage) by retaining the semantic relatedness of all different classes in the latent space via the guidance of the landmarks.
By exploring the intrinsic structure of visual data in the bottom-up projection and preserving the semantic relatedness in the top-down projection, we demonstrate that the latent representation works effectively towards bridging the semantic gap and alleviating the adversarial effect of the hubness phenomenon \citep{shigeto2015ridge}. In addition, the existing transductive post-processing techniques, e.g., \citep{fu2015transductive, zhang2016eccv}, are easily incorporated into our proposed framework to address the domain shift issue. Whenever multiple diversified visual and/or semantic representations are available, our proposed framework can further exploit the synergy among multiple representations seamlessly.

%Contributions
%\vspace{0.1cm}\noindent\textbf{Contributions}\quad
Our main contributions in this paper are summarized as follows: a) we propose a novel stagewise bidirectional latent embedding framework for zero-shot visual recognition and explore effective and efficient enabling techniques to address the semantic gap issue and to lessen the catastrophic effect of the hubness phenomenon; b) we extend our framework to scenarios in presence of multiple visual and/or different semantic representations as well as the transductive setting;  and  c) we conduct extensive experiments under a comparative study to demonstrate the effectiveness of our proposed framework on several benchmark datasets.

% paper structure
%The remainders of the paper are structured as follows.
The rest of this paper is organized as follows. Section 2 reviews related works. Section 3 presents our bidirectional latent embedding framework. Section 4 describes our experimental settings, and Section 5 reports experimental results.  The last section draws conclusions.

\section{Related Work}
\label{related}
In this section, we review existing works in zero-shot visual recognition and particularly outline connections and differences between our proposed framework and the related methods. We first provide a taxonomy on zero-shot visual recognition to facilitate our presentation and then briefly review relevant subspace learning methods that could be enabling techniques used to realize our proposed framework.

\subsection{Zero-Shot Visual Recognition}
\label{related_zsl}
There are a number of taxonomies for zero-shot visual recognition. For example, \citet{akata2016label} proposed a taxonomy that highlights two crucial choices in ZSL, i.e., the prior information and the recognition model, while the taxonomy provided by \citet{changpinyo2016synthesized} is from a perspective of knowledge transfer. To facilitate our presentation in this paper, we would divide the existing zero-shot visual recognition methods into three categories from a perspective on how the existing methods bridge the semantic gap, namely, \emph{direct mapping, model parameter transfer} and \emph{common space learning}.

%%methods of direct mapping
%\subsubsection{Direct Mapping}
\emph{Direct mapping} is a typical ZSL methodology. Its ultimate goal is learning a mapping function from visual features to semantic representations directly or indirectly \citep{lampert2009learning, lampert2014attribute, xu2015zero, xu2015semantic, jayaraman2014zero, gan2016learning, al2015transfer, akata2014zero, akata2015evaluation, akata2016label, xian2016latent, kodirov2015unsupervised, romera2015embarrassingly, shigeto2015ridge}. Such a mapping is carried out via either a classifier or a regression model depending upon an adopted semantic representation. As the relatedness between any class labels are known in semantic space or its own embedding space, a proper label may be assigned to a test instance in an unseen class by means of semantic relatedness in different manners, e.g., nearest neighbors \citep{xu2015semantic} and probabilistic  models \citep{lampert2009learning}. However, direct mapping may not be reliable in attribute predictions \citep{jayaraman2014zero, gan2016learning}. This issue has been addressed by different strategies. \citet{jayaraman2014zero} use the random forests based post-processing to handle the uncertainties of attribute predictions, while \citet{gan2016learning} propose to learn a representation transformation in visual space to enhance the attribute-level discriminative capacity for attribute prediction. Alternatively, \citet{al2015transfer} explore the additional underlying attributes by constructing the hierarchy of concepts for reliability. When the semantic representations are continuous, regression models are used to map visual features to semantic representations. A variety of loss functions along with various regularization terms have been employed to establish regression models. For example, \citet{akata2014zero}, \citet{akata2015evaluation}, \citet{akata2016label} and \citet{xian2016latent} use structured SVM to maximize the compatibility between estimated and ground-truth  semantic representations. \citet{kodirov2015unsupervised} formulate the regression as a dictionary learning and sparse coding problem. \citet{romera2015embarrassingly} make a distinction by minimising the multi-class error rather than the error of the semantic representation prediction and adding further constraints on the model parameters. In direct mapping, however, the generalization of learned mapping models is considerably limited by high intra-class variability. Furthermore, it does not address the domain shift problem well
when the training and test data are of different distributions. According to  \citet{shigeto2015ridge}, a regression model tends to project the instances closer to the origin than its ground-truth semantic representation, which exacerbates the domain shift problem.

% % model estimation
%\subsubsection{Model Parameter Transfer}
\emph{Model parameter transfer} is yet another ZSL methodology that estimates model parameters with respect to unseen classes by combining those model parameters learned from known classes via exploiting the inter-class relationship between known and unseen classes in semantic space \citep{mensink2014costa, norouzi2013zero, gan2015exploring, changpinyo2016synthesized}. Unlike direct
mapping, the zero-shot visual recognition in model parameter transfer takes place in visual space where the model parameters for unseen classes are usually obtained by a convex combination of base classifiers trained on known classes \citep{mensink2014costa, norouzi2013zero, gan2015exploring}. More recently,  \citet{changpinyo2016synthesized} proposed a novel approach that gains model parameters for unseen classes by aligning the topology of all the classes in both semantic and model parameter spaces. As a result, model parameter transfer is carried out by exploring base classifiers corresponding to ``phantom" classes, which are artificially created and not associated with any real classes, to enhance the flexibility of the model.
%Nevertheless, the hubness phenomenon could limit the performance of such methods given the fact that visual space is often of a very high dimensionality.
Since the inter-class relationship among unseen classes is not taken into account, model parameter transfer might be subject to limitation due to a lack of sufficient information for knowledge transfer.

% % common space learning
%\subsubsection{Common Space Learning}
\emph{Common space learning} is a generic methodology towards bridging the semantic gap and has been applied in ZSL \citep{fu2015transductive, zhang2015zero, zhang2016zero, changpinyo2016predicting} as well as other computer vision applications such as image retrieval \citep{gong2014multi} and automatic image description generation \citep{karpathy2015deep}. This methodology learns a common  representation space into which both visual features and semantic representations are projected for effective knowledge transfer. Consequently, zero-shot visual recognition is obtained in this learned common representation space, which is different from direct mapping, where the recognition is obtained in semantic space or its own embedding space that differs from visual embedding space in some direct mapping methods \citep{akata2015evaluation, akata2016label}, and model parameter transfer, where the recognition takes place in visual space. A learned common space may be either interpretable \citep{zhang2015zero} or latent \citep{changpinyo2016predicting, fu2015transductive, zhang2016zero}.
\citet{zhang2015zero} come up with a semantic similarity embedding method, which leads to semantic space where similarity can be readily measured for zero-shot visual recognition. This method works on viewing any instance in unseen classes as a mixture of those in known classes. More recently,  \citet{zhang2016zero} further propose a probabilistic framework for learning joint similarity latent embedding where both visual and semantic embedding along with a class-independent similarity measure are learned simultaneously. As a result, zero-shot visual recognition is obtained via optimization in the joint similarity latent space. \citet{fu2015transductive} use the \emph{canonical correlation analysis} (CCA) to project multiple views of visual data onto a common latent embedding space to address the domain shift issue.
%%In general, the learned common space tends to narrow the semantic gap and often has a lower dimension than that of visual and semantic spaces.
%Hence, the hubness phenomenon and the domain shift problem may be alleviated to some extent.
When we prepared this manuscript, one latest zero-shot recognition method \citep{changpinyo2016predicting} emerged, which involves two subsequent learning stages. Nevertheless, the generalization capability of the aforementioned common space learning models is generally limited as the intra-class variability is not tackled effectively.
\par
Our proposed framework can be viewed as a common space learning approach as zero-shot recognition is obtained in the learned common representation space (c.f. Section \ref{method}).
%In our work, we not only take visual and semantic spaces into account in the top-down stage but also effectively deal with the intra-class and the inter-class variability in the bottom-up stage.
While all common space learning methods share the same ultimate goal to bridge the semantic gap, their strategies and enabling techniques for attaining this goal may be quite different. To this end, our proposed framework consists of two subsequent learning stages, while most of other common space learning methods fulfil the joint embedding from both visual and semantic spaces simultaneously, e.g.,  \citep{fu2015transductive, zhang2015zero, zhang2016zero}. Furthermore, our framework  tackles the intra-class and inter-class variability in the common space and knowledge transfer explicitly with proper enabling techniques, while other common space learning methods address such issues implicitly, e.g., \citep{zhang2015zero, zhang2016zero} or do not take into account intra-class and inter-class variability in the latent space, e.g., \citep{changpinyo2016predicting}.
In terms of enabling techniques, other common space learning methods \citep{fu2015transductive, zhang2015zero, zhang2016zero, changpinyo2016predicting}
employ different parametric learning models for common space learning with their formulated objectives, while we address this issue by using both parametric (bottom-up) and non-parametric (top-down) learning models. The use of non-parametric model in our proposed framework allows for carrying out knowledge transfer explicitly, which readily distinguishes ours from all the existing common space learning methods that realize knowledge transfer implicitly with a parametric model that relies on the capacity in interpolation and extrapolation for generalization.

\subsection{Subspace Learning}
\label{related_subspace}
Subspace learning aims to find a low-dimensional space for high-dimensional raw data to reside in by preserving and highlighting useful information retained in the data in the high-dimensional space. In ZSL tasks, both the visual and semantic representation spaces could be of a very high dimensionality. To deal with the ``curse of dimensionality", subspace learning is often employed to address this issue in ZSL
\citep{akata2016label}. In particular, it is essential for common space learning \citep{fu2010manifold, fu2015transductive,zhang2015zero, zhang2016zero}.
In general, subspace learning models are either parametric or non-parametric.

A parametric model learns a projection from a \emph{source} high-dimensional space to a \emph{target} low-dimensional subspace via optimizing certain objectives of interest. For example, \emph{principle component analysis} (PCA) \citep{jolliffe2002principal} learns a projection that maps data points to a set of uncorrelated components accounting for as much of the variability underlying a data set as possible. \emph{Locality preserving projection} (LPP) \citep{niyogi2004locality} learns a projection for preserving the local neighborhoods in the source space. In a supervised learning scenario, a discriminative subspace can be learned by using label information. For example, \emph{linear discriminant analysis} (LDA) \citep{cai2007semi} leads to a projection that maximizes the separability of projected data points in the LDA subspace. LPP has also been extended to its supervised version by taking the label information into account \citep{cheng2005supervised}. In our work, we apply the supervised LPP algorithm as an enabling technique for learning a low-dimensional latent space from visual space.

Unlike the aforementioned parametric models, a non-parametric subspace model often learns projecting a set of high-dimensional data points onto a low-dimensional subspace directly to preserve the intrinsic properties in source space. Non-parametric models are suitable especially for a scenario that all the data points in the source space are known or available and the embedding task needs to be undertaken on a given data set without the need of extension to unseen data points during learning. This is a salient characteristic that distinguishes between parametric and non-parametric subspace learning.  As a typical non-parametric subspace learning framework,  \emph{multi-dimensional scaling} (MDS) \citep{cox2000multidimensional} refers to a family of algorithms  that learn embedding a set of given high-dimensional data points into a low-dimensional subspace by preserving the distance information between data points in the high-dimensional space. Sammon mapping \citep{sammon1969nonlinear} is an effective non-linear MDS algorithm. In our work, we extend the Sammon mapping to a semi-supervised scenario that for a given dataset the embedding of some data points in the subspace is known or fixed in advance and only remaining data points need to be embedded via preserving their distance information to others. To the best of our knowledge, this is a brand new problem that has never been considered in literature but emerges from our proposed framework for knowledge transfer between known and unseen classes.
%By using our proposed semi-supervised Sammon mapping algorithm as an enabling technique of our framework, we can embed the semantic representations of all the unseen classes onto the latent space learned from visual space with the guidance of the training-class embedding so that zero-shot visual recognition can be undertaken in the latent embedding space.

\begin{figure*}
\centering
\floatbox[{\capbeside\thisfloatsetup{capbesideposition={left,top},capbesidewidth=6cm}}]{figure}[\FBwidth]
{\includegraphics[width = 5.4in]{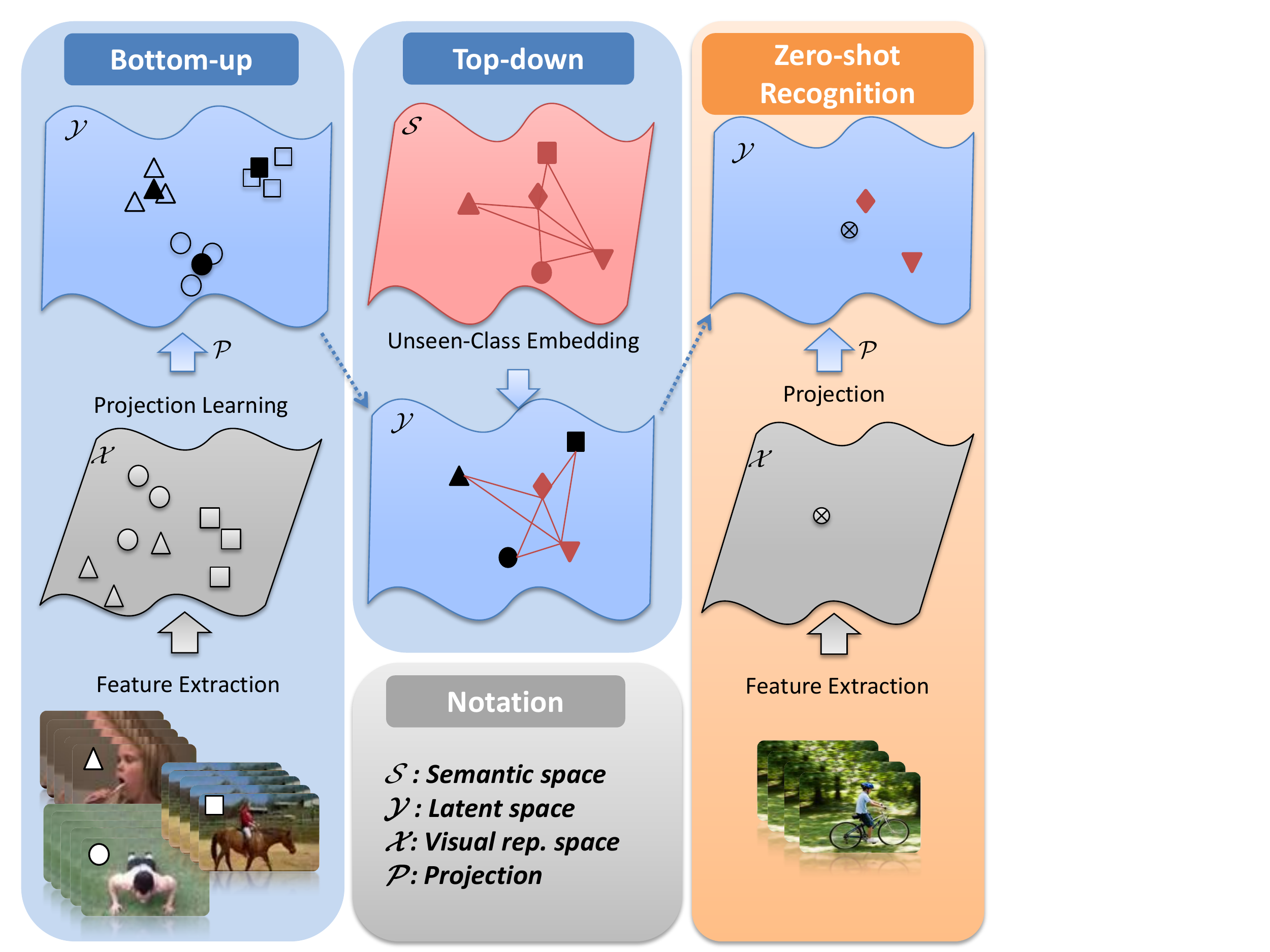}}
{\caption{The proposed bidirectional latent embedding learning (BiDiLEL) framework for zero-shot visual recognition. The BiDiLEL framework consists of two subsequent learning stages.
\newline
In the \emph{bottom-up} stage (left plot), visual representations in $\mathcal{X}$ are first extracted from the labeled visual data of different training classes  marked by $\triangle$, $\bigcirc$ and $\Box$, respectively. Then a projection $\mathcal{P}$ is learned with a proper supervised subspace learning algorithm to create a latent space $\mathcal{Y}$. The latent embedding of training-class labels are formed by using the mean of the projections of their corresponding training instances in $\mathcal{Y}$, named \emph{landmarks}, marked by $\blacktriangle$, \tikzcircle{3pt} and $\blacksquare$, respectively.
\newline
In the \emph{top-down} stage (middle plot), the unseen-class labels in the semantic space $\mathcal{S}$, marked by $\blacklozenge$ and $\blacktriangledown$, are embedded into $\mathcal{Y}$ with a landmark-based learning algorithm in order to preserve the semantic relatedness between all different classes.
\newline
For \emph{zero-shot recognition} (right plot), the visual representation of a test instance in $\mathcal{X}$, marked by $\bigotimes$, is projected into the latent space $\mathcal{Y}$ via $\mathcal{P}$ learned in the bottom-up stage. For decision-making, the nearest-neighbor rule is applied by finding out the unseen-class embedding that has the least distance to this instance in $\mathcal{Y}$. That is, the unseen-class label marked by $\blacklozenge$ is assigned to this test instance marked by $\bigotimes$.}
\label{figure_framework}}
\end{figure*}

\section{Bidirectional Latent Embedding}
\label{method}

In this section, we propose a novel framework for zero-shot visual recognition via  \emph{bidirectional latent embedding learning} (BiDiLEL). We first provide an overview on our basic ideas and the problem formulation. Then, we present the bottom-up and the top-down embedding learning with proper enabling techniques, respectively. Finally, we describe the learning model deployment for zero-shot recognition as well as two post-processing techniques for the transductive setting. To facilitate our presentation, Table \ref{table_notation} summarizes the notations used in this paper.

%The table of notations
\begin{table*}
\centering
\caption[]{Nomenclature.}
\label{table_notation}
\newsavebox{\tablebox}
\begin{lrbox}{\tablebox}
\begin{tabularx}{\textwidth}{@{}lX}\toprule
 Notation & Description \\
 \toprule
 $n_{l}$, $n_{u}$ & number of labelled (training) and unlabelled (test) instances\\
 $d_x$, $d_y$, $d_s$ & dimensionality of visual, latent and semantic spaces\\
 $X^{l} \in \mathbb{R}^{d_x\times n_{l}}$, $\pmb{x}^l_i$ &  visual representation matrix of all the labelled instances, a column corresponding to an instance \\
 $X^{u} \in \mathbb{R}^{d_x\times n_{u}}$, $\pmb{x}^u_i$ &  visual representation matrix of unlabelled instances, a column corresponding to an instance \\
 $Y^l \in \mathbb{R}^{d_y\times n_l}$, $\pmb{y}^l_i$ & projections of $X^l$ in the latent subspace $\mathcal{Y}$, a column corresponding to an instance\\
 $Y^u \in \mathbb{R}^{d_y \times n_u}$, $\pmb{y}^u_i$ & projections of $X^u$ in the latent subspace $\mathcal{Y}$, a column corresponding to an instance\\
 %$K^l \in \mathbb{R}^{n_l \times n_l}$, $K^u \in \mathbb{R}^{n_l \times n_u}$ &  representation matrices of labelled and unlabelled instances in the kernel space \\
 $W\in \mathbb{R}^{n_l \times n_l}$, $L\in \mathbb{R}^{n_l \times n_l}$ & similarity and Laplacian matrices of a given data set of $n_l$ instances\\
 $P\in \mathbb{R}^{n_l\times d_y}$ & projection matrix learned in the bottom-up stage\\
 $\mathcal{C}^l$, $\mathcal{C}^u$, $|\mathcal{C}^l|$, $|\mathcal{C}^u|$ &  known and unseen class label sets and the number of known and unseen classes in two sets\\
 $B^l\in \mathbb{R}^{d_y \times |\mathcal{C}^l|}$,$\pmb{b}^l_i$ & latent embedding for known class labels, a column corresponding to one class\\
 $B^u\in \mathbb{R}^{d_y \times |\mathcal{C}^u|}$,$\pmb{b}^u_i$ &  latent embedding for unseen class labels learned in the top-down stage, a column corresponding to one class\\
\bottomrule
\end{tabularx}
\end{lrbox}
\scalebox{1}{\usebox{\tablebox}}
\end{table*}

\subsection{Overview}
\label{overview}
%overview and motivations

%It is well known that the semantic gap is a main hurdle that greatly hinders progresses in zero-shot visual recognition. To tackle the semantic gap challenge, we propose a framework named bidirectional latent embedding.

The motivation behind our proposed framework is two-fold: a) to narrow the semantic gap, a latent space is learned from visual representations of training data in a supervised manner by preserving
intrinsic structures underlying visual data and promoting the discriminative capability simultaneously and b) for knowledge transfer, the semantic representations of unseen-class labels are then embedded into the learned latent space of favorable properties by taking into account both the embedding of training-class labels and the semantic relatedness between all different classes; i.e., not only the relationships between known and unseen classes but also that between unseen classes.
Based on our motivation described above, we propose a framework of a sequential bidirectional learning strategy: the bottom-up learning for creating the latent space from visual data and then the top-down learning for embedding all the unseen-class labels in the learned latent space, as illustrated in Fig. ~\ref{figure_framework}.

In the bottom-up stage, the visual representations of training examples are extracted.  A proper supervised subspace learning algorithm is employed to learn a projection $\mathcal{P}$ for preserving the intrinsic locality of instances within the same class and promoting the separability of instances in different classes. As a result, a discriminative latent space $\mathcal{Y}$ is created.  Then, we estimate the mean of projections of training instances for every training class. All the estimated means of training classes in $\mathcal{Y}$ are designated for their latent embedding of training-class labels specified in $\mathcal{C}^l$. As a result, we expect that the the bottom-up learning creates the latent embedding of training-class labels that better reflects the semantic relatedness among them and lowers the intra-class variability simultaneously. Thus, we designate all the estimated means of training classes as \emph{landmarks} in the latent space and would use them to guide the embedding of unseen-class labels specified in $\mathcal{C}^u$ into the same latent space. The bottom-up latent space learning is carried out by a supervised subspace learning algorithm, \emph{supervised locality preserving projection} (SLPP) \citep{cheng2005supervised}, which is presented in Section \ref{sub:bottom-up}. The motivation behind this choice is to deal with intra-class and inter-class issues along with preserving the intrinsic structure underlying visual data. \emph{Locality preserving projection} (LPP) \citep{niyogi2004locality} is an algorithm that preserves intrinsic structure underly data, as shown in \citep{ niyogi2004locality}. Its supervised version, SLPP, further exploits the labeling information to lower the intra-class variability and hence improves the separability between different classes, as shown in \citep{cheng2005supervised, zhang2010locality, zheng2007gabor}.

As no training examples in unseen classes are available in ZSL, we have no information on their properties in visual space but clearly know the semantic relatedness between different class labels by means of their semantic representations. In the top-down stage, we thus embed unseen-class labels into the latent space by preserving the semantic relatedness between all different class labels, including training-class to unseen-class as well as
unseen-class to unseen-class, guided by the landmarks. Such top-down learning requires a proper enabling technique. To the best of our knowledge, no existing algorithm meets this requirement. Therefore, we propose a semi-supervised MDS algorithm based on the Sammon mapping \citep{sammon1969nonlinear}, named \emph{landmark-based Sammon mapping} (LSM), as our enabling technique to learn the latent embedding of unseen-class labels, which is presented in Section \ref{sub:top-down}.

Once the two subsequent learning tasks are carried out, zero-shot visual recognition is easily obtained in the latent space with a nearest-neighbor rule presented in Section \ref{sub:recognition}.

Now, we formulate the general problem statement for zero-shot visual recognition. Given a set of labelled instances $X^{l} = \{\pmb{x}^l_1, \pmb{x}^l_2,... , \pmb{x}^l_{n_{l}}\} \in \mathcal{X}, ~\pmb{x}_i \in \mathbb{R}^{d_x}$, their labels are denoted by $Z^l = \{z^l_1, z^l_2, ... , z^l_{n_l}\}, ~z^l_i \in \mathcal{C}^l$, where $\mathcal{C}^l$ is the set of known class labels. For any given unlabelled instance set $X^{u} \in \mathbb{R}^{d_x \times n_{u}}$, the zero-shot visual recognition problem is to predict their labels in $\mathcal{C}^u$ that properly describe the test instances by assuming $\{z^u_i\} \in \mathcal{C}^u$  and $\mathcal{C}^l \cap \mathcal{C}^u = \emptyset$. Here, $n_{l}$ and $n_{u}$ are the number of labelled (training) and unlabelled (test) instances, respectively, and $d_x$ is the dimensionality of a visual representation.

\subsection{Bottom-up Latent Space Learning}
\label{sub:bottom-up}

The bottom-up latent space learning aims to find a projection matrix $P$ that maps instances from their visual space $\mathcal{X}$ to a latent space of a lower dimension $\mathcal{Y}$ to preserve the intrinsic locality of instances within the same class and to promote the separability of instances in different classes. While there are a number of candidate techniques to learn such a latent space, we employ the (SLPP) \citep{cheng2005supervised} as the enabling technique since it generally outperforms other candidate techniques,
%in terms of preserving the local structures in visual space and improving the discriminative capability in the learned latent space,
as validated in Section \ref{results}.

In SLPP, a graph is first constructed with all the training data in $X^{l}$ to characterize the manifold underlying this data set in the visual representation space $\mathcal{X}$. Following the original settings used in the LPP algorithm \citep{niyogi2004locality}, $k$ nearest neighbors ($k$NN) of a specific data point are used to specify its neighborhood for the graph construction. Training instances $\pmb{x}_i^l \in X^{l}$ are represented by the nodes in the graph, and an edge is employed to link two nodes when one  is in the other's $k$NN neighborhood. Unlike the unsupervised LPP algorithm, we further take into account  the labelling information of the instances when constructing the graph  \citep{cheng2005supervised}. As a result, the edge between two nodes is removed when they do not share the same class label. Therefore, we have a similarity matrix containing all the weights of edges as follows:
\begin{equation}
\label{eq2}
W_{ij} = \left \{
\begin{array}{ll}
\exp(-||\mathbf{x}_i^l-\mathbf{x}_j^l||/2), & \mathbf{x}_i^l \in \mathcal{N}_k (\mathbf{x}_j^l)~ \textrm{or} ~~\mathbf{x}_j^l \in \mathcal{N}_k (\mathbf{x}_i^l),\\
   &z_i^l=z_j^l\\
0 ,&  otherwise
\end{array}
\right.
\end{equation}
where $\mathcal{N}_k(\pmb{x})$ denotes the set of $k$ nearest neighbours of $\pmb{x}$.

In order to preserve the intrinsic local structure, we use the following cost function for learning a projection $P$:
\begin{equation}
\label{eq1}
L(P; W, X^l) = \sum_{i,j} ||P^T \pmb{x}^l_i-P^T \pmb{x}^l_j||_2^2 W_{ij},
\end{equation}
where $\pmb{x}^l_i$ is the $i$-th column of the input data matrix $X^l$, corresponding to the feature vector of the $i$-th training example.

Minimizing the cost function in Eq.(\ref{eq1}) enables the nearby instances of the same class label in the visual space to stay as close as possible in the learned latent space. Hence, the intra-class variability is decreased and the inter-class variability is increased reciprocally. For the sake of robustness in numerical computation, the above optimization problem is converted into the following form with the mathematical treatment \citep{niyogi2004locality}:
\begin{equation}
\label{eq3}
\max_{P} \frac{Tr(P^TX^lD {X^l}^T P)}{Tr(P^TX^l L {X^l}^T P)},
\end{equation}
where $L=D-W$ is the laplacian matrix and $D$ is a diagonal matrix with $D_{ii}=\sum_j W_{ij}$.

To penalize the extreme values in the projection matrix $P$, we further employ a regularization term $Tr(P^TP)$. Thus the cost function in Eq. (\ref{eq1}) is now in the following form:
\begin{equation}
\label{eqObj}
\max_{P} \frac{Tr(P^TX^lD {X^l}^T P)}{Tr(P^T(X^lL {X^l}^T +\alpha I)P)}
\end{equation}
Finding the optimal projection $P$ is simply boiled down to solving the generalized eigenvalue problem:
\begin{equation}
\label{eqSol}
 X^lD{X^l}^T\pmb{p} = \lambda (X^lL{X^l}^T+\alpha I) \pmb{p},
\end{equation}
and the analytic solution is obtained by setting $P = [\pmb{p}_1,...,\pmb{p}_d]$ where $\pmb{p}_1,...,\pmb{p}_d$ are those eigenvectors corresponding to the largest $d$ eigenvalues.

Motivated by the treatment proposed by \citet{akata2013label, akata2016label} for binary label embedding, we further apply two normalization strategies, \emph{centralization} and $l_2$\emph{-normalization}, to the latent representations of training examples, $Y^l$, to avoid unfavorable situations in zero-shot  recognition. Our motivation behind the treatment is different from theirs \citep{akata2013label, akata2016label}. For the sake of readability, we have to describe our motivation at the end of Section \ref{sub:top-down} as it concerns not only bottom-up but also top-down learning stages.
By using the centralization, the latent representations $Y^l$ are centralized to make all the features (i.e., rows) have zero mean. Furthermore, $l_2$-normalization is applied on each column of $Y^l$ to make all the instances have unit norms, i.e., $\hat{\pmb{y}}^l_i = \pmb{y}^l_i/||\pmb{y}^l_i||_2$ for $i = 1,2,...,n_l$. After the centralization and $l_2$-normalization, the latent embedding of $i$-th training class, $\pmb{b}^l_i$, is estimated by
\begin{equation}
\label{eqlatEmb}
\pmb{b}^l_i = \frac{1}{n_i} \sum_{z^l_j = i} \hat{\pmb{y}}^l_j, ~~i = 1, \cdots, |\mathcal{C}^l|,
\end{equation}
where $n_i$ is the number of training instances in the $i$-th training class, and
$|\mathcal{C}^l|$ is the number of training classes. Likewise, all mean points of $|\mathcal{C}^l|$ known classes estimated from training instances, $\pmb{b}^l_1, \cdots, \pmb{b}^l_{|\mathcal{C}^l|}$, are  $l_2$-normalized to have unit norms. We specify all $|\mathcal{C}^l|$ normalized mean points as \emph{landmarks} to provide the guidance for embedding unseen classes into the learned latent space (c.f. Section \ref{sub:top-down}).

\subsection{Top-down Latent Embedding learning}
\label{sub:top-down}

The top-down algorithm aims to learn latent embedding of unseen classes. With the guidance of landmarks, i.e., the latent embedding of known classes, all the unseen-class labels are embedded into the same latent space learned in the bottom-up stage via preserving their semantic relatedness pre-defined by an existing semantic representation of class labels (c.f. Section \ref{sub:semantic}).

Let  $B^{l} = \{\pmb{b}^l_1, \pmb{b}^l_2,...,\pmb{b}^l_{|\mathcal{C}^l|}\} \in \mathbb{R}^{d_y \times |\mathcal{C}^l|}$ collectively denote the latent embedding of all the training classes where $d_y$ is the dimension of the latent space formed in the bottom-up stage. Similarly, the latent embedding of $|\mathcal{C}^u|$ unseen classes are collectively denoted by $B^{u} = \{\pmb{b}^u_1, \pmb{b}^u_2,...,\pmb{b}^u_{|\mathcal{C}^u|}\} \in \mathbb{R}^{ d_y \times |\mathcal{C}^u|}$. In order to preserve the semantic relatedness between all the classes, the distance between two classes in the latent space should be as close to their semantic distance in the semantic space as possible but the embedding of known classes are already settled with Eq. (\ref{eqlatEmb}) in the bottom-up learning stage. Hence, this leads to a brand new semi-supervised MDS problem. By means of the Sammon mapping \citep{sammon1969nonlinear}, we propose a \emph{landmark-based Sammon mapping} (LSM) algorithm to tackle this problem.

By using a proper semantic representation of all class labels, we achieve the semantic representations of training and unseen classes, $S^{l} \in \mathbb{R}^{d_s \times |C^l|}$ and $S^{u} \in \mathbb{R}^{d_s \times |C^u|}$,  where their $i$-th columns are $\pmb{s}^l_i$ and $\pmb{s}^u_i$, respectively, and $d_s$ is the dimensionality of the semantic space. Then, the LSM cost function is defined by
\begin{equation}
\label{eq6}
\begin{aligned}
&E(B^{u}) = \frac{1}{|\mathcal{C}^l| |\mathcal{C}^u|} \sum_{i=1}^{|\mathcal{C}^l|} \sum_{j=1}^{|\mathcal{C}^u|} \frac{ (d(\pmb{b}^{l}_i,\pmb{b}^{u}_j)-\delta(\pmb{s}^{l}_i,\pmb{s}^{u}_j))^2}{\delta(\pmb{s}^{l}_i,\pmb{s}^{u}_j)}\\
&+\frac{2}{|\mathcal{C}^u|(|\mathcal{C}^u|-1)} \sum_{i=1}^{|\mathcal{C}^u|} \sum_{j=i+1}^{|\mathcal{C}^u|} \frac{ (d(\pmb{b}^{u}_i,\pmb{b}^{u}_j)-\delta(\pmb{s}^{u}_i,\pmb{s}^{u}_j))^2}{\delta(\pmb{s}^{u}_i,\pmb{s}^{u}_j)},
\end{aligned}
\end{equation}
where $d(\pmb{x},\pmb{y})$ and $\delta(\pmb{x},\pmb{y})$ are the distance metrics in the latent space and the semantic space, respectively.
Intuitively, the first term of Eq. (\ref{eq6}) concerns the semantic relatedness between known and unseen classes and the second term of Eq. (\ref{eq6}) takes into account the semantic relatedness between unseen classes in the top-down learning. Minimizing  $E(B^{u})$ leads to the solution: $B^u{^*} = \arg \min_{B^u} E(B^u)$.

\begin{algorithm}[tb]
 \caption{Landmark-based Sammon Mapping (LSM)}
 \label{alg1}
 \algnewcommand\algorithmicreturn{\textbf{Return:}}
 \algnewcommand\RETURN{\item[\algorithmicreturn]}
 \renewcommand{\algorithmicrequire}{\textbf{Input:}}
 \renewcommand{\algorithmicensure}{\textbf{Output:}}
 \begin{algorithmic}[1]
 \REQUIRE The semantic representations for training and unseen classes, $S^l$ and $S^u$, (or the semantic distance matrix $\Delta=\{\delta_{ij}(\pmb{s}_i,\pmb{s}_j)\}$), the training-class latent embedding $B^l$, learning rate $\eta$.
 \ENSURE The latent unseen-class embedding ${B^u}^*$.
 \STATE Initialize $B^u_0$ for $t=0$ randomly;
 \REPEAT
 \STATE Calculate gradient $g_t = \nabla_{B^u_t} E(B^u_t)$ (c.f. Appendix \ref{appA});
 \STATE Update $B^u_{t+1} := B^u_t + \eta g_t$;
 \STATE $t := t+1$;
 \UNTIL Stopping criteria are satisfied.
 \end{algorithmic}
\end{algorithm}

Following \citet{sammon1969nonlinear}, we derive the LSM algorithm by using the gradient descent optimization procedure. As a result, our LSM algorithm is summarized in Algorithm \ref{alg1}, and the derivation of gradient $\nabla_{B^u} E(B^u)$ used in Algorithm \ref{alg1} is described in Appendix \ref{appA}. Applying Algorithm \ref{alg1} to the semantic representations of $|\mathcal{C}^u|$ unseen classes results in their embedding in the latent space:
$\pmb{b}^u_1, \cdots, \pmb{b}^u_{|\mathcal{C}^u|}$.

Now we described our motivation underlying two normalization strategies presented at the end of Section \ref{sub:bottom-up}. In general, our motivation underlying two normalization strategies aims to facilitate the embedding of unseen-class labels in the top-down stage. As advocated by \citep{akata2016label}, the instance-level $l_2$-normalization of binary attributes of class labels to the unit magnitude and zero-mean centering facilitate zero-shot recognition.  For embedding unseen classes in the latent space, our LSM algorithm has to take into account the distance information between known and unseen classes in both the semantic and the latent spaces. Applying the $l_2$-normalization to the embedding of training instances thus ensures that the distances measured in two spaces are in the same scale. Applying the centralization is due to the $l_2$-normalization. All the $l_2$-normalized training instances in the latent space may concentrate in a small region (on the one surface side of the unit hyper-sphere). This phenomenon may cause no sufficient room or a difficulty to accommodate the embedding of unseen-class labels in the top-down learning. The zero-mean centralization ameliorates the detrimental effect of this phenomenon by scattering training instances in a larger region to facilitate the unseen class label embedding.

\subsection{Zero-Shot Recognition in the Latent Space}
\label{sub:recognition}

Once all the class labels are embedded in the latent space by our Algorithm \ref{alg1}, zero-shot visual recognition is gained in the learned latent space. Given a test instance $\pmb{x}_i^u$, its label is predicted in the latent space via the following procedure. First of all, we apply projection $P$ obtained in the bottom-up learning stage to map it into the latent space:
\begin{equation}
\pmb{y}_i^u = P^T \pmb{x}_i^u.
\end{equation}
After being subtracted by the mean estimated on all the training instances in the latent space, $\pmb{y}_i^u$ is then $l_2$-normalized in the same manner as done for all training instances. Thus, its label, $l^*$, is assigned to the class label of which embedding is closest to $\pmb{y}_i^u$; i.e.,
\begin{equation}
l^* = \arg \min_l d(\pmb{y}_i^u,\pmb{b}^u_l),
\end{equation}
where $\pmb{b}^u_l$ is the latent embedding of $l$-th unseen class, and $d(\pmb{x},\pmb{y})$ is a distance metric in the latent space. In our experiments, the Euclidean distance metric is used for measuring the distance due to the nature of manifold learning in the LPP algorithm \citep{niyogi2004locality}.

A recent study \citep{shao2016kernelized} suggests that the use of multiple visual representations can improve the robustness in action recognition. As a result, we have extended our proposed framework to the joint use of multiple complimentary visual representations
for robust zero-shot visual recognition, which is presented in Appendix \ref{appB}. To promote robustness, we also come up with a visual representation complementarity measurement, as described in Appendix \ref{appC}.

\subsection{Post-processing Techniques}
\label{selftraining}

The post-processing in ZSL refers to those techniques that exploit the information conveyed in test instances to improve the ZSL performance. In our work, two existing post-processing techniques,
\emph{self-training} \citet{xu2015zero} and \emph{structured prediction} \citep{zhang2016eccv}, are incorporated into our proposed framework.

\subsubsection{Self-training}
The self-training (ST) is a post-processing technique proposed by \citet{xu2015zero} in order to alleviate the domain shift problem. The general idea behind the self-training is adjusting the latent embedding of unseen classes according to the distribution of all the test instance projections in the latent space. It is straightforward to incorporate this post-processing technique into our zero-short visual recognition framework. Given the $i$-th unseen class ($i= 1,2,...,|\mathcal{C}^u|$),  \citet{xu2015zero} adjust the latent embedding $\pmb{b}_i^u$  to $\hat{\pmb{b}}_i^u$, where
\begin{equation}
\label{eq13}
\hat{\pmb{b}}_i^u := \frac{1}{k} \sum_{\pmb{y}^u \in \mathcal{N}_{k}(\pmb{b}_i^u)}^{k} \pmb{y}^u.
\end{equation}
Here, $\mathcal{N}_k(\pmb{b}_i^u)$ is a neighborhood of the latent embedding $\pmb{b}_i^u$ containing the $k$ nearest test instances. In other words, this nearest neighbour search in the self-training is confined to only test instances. As all the test instances have to be used in the self-training, this leads to a\emph{ transductive} learning setting. Unlike their treatment in \citep{xu2015zero}, in our experiments, we adjust $\pmb{b}_i^u$ to the arithmetic average between $\hat{\pmb{b}}_i^u$ and $\pmb{b}_i^u$,  $(\hat{\pmb{b}}_i^u+\pmb{b}_i^u)/2$, for a trade-off between preserving their semantic relatedness and alleviating the domain shift effect.

\subsubsection{Structured Prediction}
Structured prediction is yet another option for post-processing recently proposed by \citet{zhang2016eccv}. Similar to self-training, structured prediction also takes advantage of the batch of test instances under the transductive setting. This method was originally proposed for their own zero-shot recognition algorithm \citep{zhang2016zero}. In our work, we adapt it for our proposed framework, which is a simplified version of their structured prediction algorithm \citep{zhang2016eccv} by using only its first step and dropping out the rest steps due to incompatibility to our approach.

In this simplified version, we update the latent embedding of unseen classes $B^{u}$ by clustering analysis on the batch of test instances. First of all, a number of clusters are generated for all the test instances by the \emph{K}-means algorithm where the number of clusters is chosen the same as that of unseen classes $|\mathcal{C}^u|$. In our experiments, we always initialize the cluster centers with the latent embedding of unseen-class labels learned in the top-down stage\footnote{Our empirical study suggests that the random initialization in the $K$-mean clustering may lead to better performance but causes structured prediction to be unstable.}.  After the $K$-mean clustering, structured prediction needs to establish a one-to-one correspondence between a cluster and a unseen class so that the sum of distances of all possible pairs of cluster center and the unseen-class embedding can be least. Let $A \in \{0,1\}^{|\mathcal{C}^u| \times |\mathcal{C}^u|}$  denote the one-to-one correspondence matrix where $A_{ij}=1$ indicates that cluster $i$ corresponds to unseen class $j$. The correspondence problem is formally formulated as follows:
\begin{eqnarray}
\label{eq:strucPredict}
 \min_{A} \sum_{c = 1}^{|\mathcal{C}^u|} \sum_{k = 1}^{|\mathcal{C}^u|} A_{kc}\cdot d(\pmb{m}_k, && \pmb{b}^u_c) \nonumber \\
&&
\hspace*{-2.5cm}
\quad s.t. \quad \forall k, \forall c,~~ \sum_k A_{kc} = 1, \sum_c A_{kc} = 1,
\end{eqnarray}
where $\pmb{m}_k$ is the center of $k$-th cluster, $\pmb{b}_c^u$ is the $c$-th unseen-class latent embedding and $d(\cdot , \cdot)$ is Euclidean distance metric. This optimization problem in Eq. (\ref{eq:strucPredict}) can be solved by linear programming \citep{zhang2016eccv}.

For zero-shot recognition, a test instance falling into a specific cluster is assigned to the label of its corresponding unseen class based on the correspondence matrix $A$.

\section{Experimental Settings}
\label{experiment}

In this section, we describe our experimental settings including the information of benchmark datasets, the visual and the semantic representations used in our experiments, the investigation of different factors that may affect the zero-shot visual recognition accuracy and our comparative study.

\subsection{Dataset}
\label{dataset}

In our experiments, we employ four publicly accessible datasets to evaluate our proposed framework. The first two are benchmarks for zero-shot object recognition, namely animal with attributes (AwA) \citep{lampert2014attribute} and Caltech-UCSD Birds-200-2011 (CUB-200-2011) \citep{wah2011caltech}. As both are among those most commonly used datasets used to evaluate ZSL algorithms in literature, we can directly compare the performance of our approach  to that of those state-of-the-art zero-shot visual recognition methods. Other two datasets are UCF101 \citep{soomro2012ucf101} and HMDB51 \citep{kuehne2011hmdb}, which are benchmarks widely used to evaluate the performance of a human action recognition algorithm in presence of a large number of classes. To evaluate the performance in zero-shot human action recognition, we use the same class-wise data splits on UCF101 and HMDB51 as suggested by \citet{xu2015semantic, xu2015zero} in our experiments, which allows us to compare ours to theirs explicitly.

\begin{table}[tb]
\label{tab:dataset}
\centering
\caption[]{Summary of datasets used in our experiments}
\label{table_summary}
%\newsavebox{\tablebox}
\begin{lrbox}{\tablebox}
\begin{tabular}{@{}ccccc}\toprule
 \textbf{Number}  & \textbf{AwA} & \textbf{CUB-200-2011} & \textbf{UCF101} & \textbf{HMDB51} \\
 \toprule
 Attributes & 85 & 312& 115 & -  \\
 Known classes & 40 & 150& 51/81 & 26 \\
 Unseen classes   & 10 & 50& 50/20 & 25 \\
 Instances   & 30,475 & 11,788& 13,320 & 6,676 \\
\bottomrule
\end{tabular}
\end{lrbox}
\scalebox{0.9}{\usebox{\tablebox}}
\end{table}

Table \ref{table_summary} summarizes the main information of four datasets used in our experiments. The specific setting for zero-shot visual recognition is highlighted as follows:
%%%%
\begin{easylist}[itemize]
& \textbf{AwA}: there are 30,475 animal images belonging to 50 classes. The 40/10 (known/unseen) class-wise data split has been originally set by the dataset collectors \citep{lampert2014attribute}.
& \textbf{CUB-200-2011}: this is a fine-grained dataset of 11,788 images regarding 200 different bird species, collected by \citet{wah2011caltech}.
The class-wise data split is often 150/50 (known/unseen) on this dataset in previous works.
In our experiments, we follow the same 100/50/50 class-wise data split for training/validation/test used in \citep{akata2015evaluation, reed2016learning, xian2016latent}.
& \textbf{UCF101}: it is a human action recognition dataset collected from YouTube by \citet{soomro2012ucf101}. There are 13,320 real action video clips falling into 101 action categories. In our experiments, we use 51/50 and 81/20 (known/unseen) class-wise data splits. We use the same 30 independent 51/50 splits\footnote{The dataset of all 30 splits are available online: http://www.eecs.qmul.ac.uk/~xx302/.} randomly generated by \citet{xu2015semantic}. Regarding 81/20 splits, we randomly generate 30 independent splits as this setting does not appear in their work  \citep{xu2015semantic}.
& \textbf{HMDB51}: it contains 6,766 video clips from 51 human action classes, collected by \citet{kuehne2011hmdb}. Once again, we use the same 30 independent 26/25 splits randomly generated by \citet{xu2015semantic}.
\end{easylist}

\subsection{Visual Representation}

The latest progresses in computer vision suggest that features learned by using deep \emph{convolutional neural networks} (CNNs) significantly outperform any of hand-crafted counterparts in object recognition \citep{simonyan2014very, szegedy2015going}. Features learned by deep CNNs have also been applied in zero-shot visual recognition  \citep{akata2014zero, al2015transfer, fu2015transductive}. In our experiments, we use two different pre-trained deep CNN models to generate visual representations of images in AwA and CUB-200-2011. For a direct comparison with state-of-the-art methods, we follow their settings by using the top fully connected layer of GoogLeNet of 1024 dimensions \citep{szegedy2015going} and the top pooling layer of VGG19 of 4096 dimensions \citep{simonyan2014very} to generate feature vectors of images. In particular, MatConvNet \citep{vedaldi15matconvnet} has been employed to extract the aforementioned deep features.

There are many different visual representations that characterize video streams regarding human actions. After investigating the existing visual representations for human action video streams, we employ two kinds of state-of-the-art visual representations for human action video streams in our experiments, i.e. the \emph{improved dense trajectory} (IDT) \citep{wang2013action} and the \emph{convolutional 3D} (C3D) \citep{tran2014learning}. Our empirical studies described in Appendix \ref{appC} along with those reported in literature suggest that two selected visual representations not only outperform a number of candidate representations but also are highly complementary to each other.
The IDT is a class of state-of-the-art hand-crafted visual representations proposed by \citet{wang2013action} for human action recognition. Four different types of visual descriptors, HOG, HOF, MBHx and MBHy, are extracted from each spatio-temporal volume, and their dimensions are reduced by a factor of two with PCA. Then the representations of a video stream are generated by the Fisher vector derived from a Gaussian mixture model of 256 components. Thus, the video representations have  24,576 features for HOG, MBHx, MBHy and 27,648 for HOF \citep{wang2013action, peng2014bag}, respectively.
For computational efficiency, we further apply PCA on those video representations to reduce their dimensions down to 3,000 in our experiments.
Note that the visual representation, IDT(MBH), in our experiments refers to a feature vector formed by concatenating MBHx and MBHy.
C3D \citep{tran2014learning} is an effective approach that uses deep CNNs for spatio-temporal video representation learning. In our experiments, we use the model provided by \citet{tran2014learning}. This model was pre-trained on the Sports-1M dataset. Following the settings in \citep{tran2014learning}, we divide a video stream into segments in length of 16 frames and there is an overlap of eight frames on two consecutive segments. As a result, the fc6 activations are first extracted for all the segments and then averaged to form a 4096-dimensional video representation.

In our experiments for multiple visual representations, different visual representations described above are jointly used via our proposed combination approach described in Appendix \ref{appB}.

\subsection{Semantic Representation}
\label{sub:semantic}

To evaluate our proposed framework thoroughly, we employ two widely used semantic representations,  \emph{attributes} and \emph{word vectors}, in our experiments.

As shown in Table \ref{table_summary}, AwA and CUB-200-2011 self-contain 85 and 312 class-level continuous attributes that characterize each class label, respectively. UCF101 class labels have been manually annotated with 115 binary attributes by \citet{THUMOS14}. To our knowledge, however, there are no attributes for those class labels appearing in HMDB51. Hence, we cannot report attribute-based results on this dataset. Table \ref{table_att} exemplifies some typical attributes used in different datasets. Following the suggestion made by \citet{akata2016label}, \citet{changpinyo2016synthesized} and \citet{zhang2015zero}, we also apply $l_2$-normalization to each of attributes vectors to facilitate their latent embedding. In our experiments, we use Euclidean distance metric to measure the semantic distance between attributes of two class labels during the top-down latent embedding learning.

\begin{table}
\centering
\caption[]{Exemplification of typical attributes used in different datasets.}
\label{table_att}
%\newsavebox{\tablebox}
\begin{lrbox}{\tablebox}
\begin{tabular}{p{1.5cm}p{6.5cm}}\toprule
 \textbf{Dataset}  & \textbf{Attribute} \\
 \toprule
 AwA & colours(black, brown, red, etc.), stripes, furry, hairless,
 big, small, paws, longneck, tail, chewteeth, fast, smelly,
 bipedal, jungle, water, cave, group, grazer, insects\\ \midrule
 CUB-200-2011 & bill\_shape(curved, dagger, hooked, needle, etc.),
 wing\_color(blue, yellow, etc.), upperparts\_color,
 tail\_shape(forked, rounded, pointed, squared, etc.) \\\midrule
 UCF101   & object(ball\_like, rope\_like, animal, sharp, etc.),
 bodyparts\_visible(face, fullbody, onehand, etc.),
 body\_motion(flipping, walking, diving, bending, etc.)
  \\
\bottomrule
\end{tabular}
\end{lrbox}
\scalebox{0.9}{\usebox{\tablebox}}
\end{table}

Unlike attribute-based semantic representations, \citet{mikolov2013distributed} propose a continuous skip-gram model to learn a distributed semantic representation, \emph{word vectors}, in an unsupervised way. In our experiments, we employ the skip-gram model (well known as \emph{Word2Vec}) \citep{mikolov2013distributed}, trained on the Google News dataset containing about 100 billion words for AwA, UCF101 and HMDB51, where the word embedding space is of 300 dimensions. However, there are a number of out-of-vocabulary words in CUB-200-2011. As a result, we employ 400-dimensional word vectors trained on English-language Wikipedia \citep{akata2015evaluation, xian2016latent} for CUB-200-2011. Following the existing works, we use the ``cosine" distance metric to measure the semantic distance between two class labels in a word embedding space during the top-down latent embedding learning.

\subsection{On Hyper-Parameters}
\label{sub:hyper}

It is well known that hyper-parameters in a learning model may critically determine its performance. Thus, we investigate the impact of different hyper-parameters involved in our proposed framework to search for ``optimal" hyper-parameter values. In general, there are four hyper-parameters; i.e.,  the number of nearest neighbors ($k_G$) for the graph construction in SLPP, the trade-off factor ($\alpha$) applied to the regularization in SLPP and the dimensionality of a learned latent space ($d_y$) during the bottom-up latent embedding learning as well as the number of nearest neighbors ($k_{ST}$) when the self-training \citep{xu2015zero} is used.

In our experiments, we use the \emph{classwise} cross-validation to seek the optimal hyper-parameter values and investigate how each hyper-parameter affects the performance. We strictly follow the procedure suggested  by \citet{akata2016label,zhang2016zero} to do the cross-validation on all the datasets apart from CUB-200-2011 that has a standard training/validataion/test split. In a trial, we randomly reserve 20\% training classes as validation data and the rest of training classes are used as training data. In our experiments, we repeat such a cross-validation experiment for multiple trials and report the averaging performance on validation data. For AwA, five trials were conducted in our cross-validation based on its default training/test split. For two human action datasets, UCF101 and HMDB51, each has 30 different training/test splits provided by \citet{xu2015semantic}.  For each of 30 splits, we conducted three-trial cross-validation to achieve the optimal hyper-parameter values for this split only. Hence, our cross-validation experiment on a human action dataset had to be repeated for 30 times on all the splits respectively.

Without considering the post-processing of self-training, our approach has three hyper-parameters, $\alpha$, $d_y$ and $k_G$. It would be extremely expensive computationally if an exhausted grid search is conducted. In our experiments, we adopt a two-stage procedure to find out optimal hyper-parameters for different visual representations respectively. We first conducted a coarse grid search with $\alpha = 0.1, 10$, $d_y = 10, 100, 500$, and $k_G = 1, 10, 50$. Then, we further fine-tune each of hyper-parameters sequentially by fixing the remaining two hyper-parameters.

In our fine-tuning stage, we conduct the cross-validation experiments for each of four hyper-parameters sequentially 
based on the information (on how sensitive a hyper-parameter is to the performance) obtained from the coarse grid search. Thus, our fine-tuning stage performs in the following order:

\begin{easylist}[itemize]
& $\pmb \alpha$: First of all, we investigate the impact of $\alpha$ in Eq.(\ref{eqObj}). In our experiment, we fix the initial optimal value of $d_y$ and $k_G$
%$d_y = 100$ and $k_G = 10$
resulting from the grid search to look into the impact of $\alpha$ by setting it to $0.001, 0.01, 0.1, 1, 10, 100$ and $1000$.
& $\pmb d_y$:  As training class labels are used in the bottom-up latent embedding learning, the proper value of $d_y$ may depend on the number of training classes that varies across different datasets. To investigate the zero-shot recognition accuracy with different $d_y$ values in a large range, we use the optimal values of $\alpha$ found in the previous step and fix the initial optimal value $k_G$
 %$k_G=10$
resulting from the grid search. In our experiment, we look into $d_y = 50, 100, 150, 200, 250$ and $300$.
& $\pmb k_G$: By making use of the optimal $\alpha$ and $d_y$ values achieved from two previous steps, we look into the impact of $k_G$ defined in Eq.(\ref{eq2}) for each dataset in the same manner by fixing other hyper-parameters and allowing only $k_G$ to change in a large range: $k_G = 5, 10, 15, 20, 25$ and $30$, respectively, to see how $k_G$ affects the zero-shot recognition accuracy on different datasets.
&  $\pmb k_{ST}$: For this post-processing, we fix the optimal values of three hyper-parameters found as described above and evaluate the zero-shot recognition accuracy with a large range of $k_{ST}$ in Eq.(\ref{eq13}) from 20 to 200 with an interval of 20 on each dataset, as suggested in \citet{xu2015zero}.
\end{easylist}

As a result, the set of hyper-parameter values leading to the \emph{best} accuracy in the above fine-tuning process are treated as \emph{``optimal"} and used in test to yield the performance for unseen classes.

\subsection{On Enabling Techniques}
\label{sub:subspace}

This experimental setting aims to explore the proper enabling techniques for our proposed framework and investigate the role played by two subsequent learning stages. As stated in Section \ref{overview}, there are a number of candidate subspace learning techniques that could be used in the bottom-up learning as reviewed in Section \ref{related_subspace}. To the best of our knowledge, however, none of the existing non-parametric subspace learning model can be directly applied to the top-down learning where the task emerges from our proposed framework (c.f. Section \ref{sub:top-down}). Motivated by the work \citep{changpinyo2016predicting}, we employ a parametric learning model as a baseline for the top-down learning. In all the experiments described below, the nearest-neighbor rule described in Section \ref{sub:recognition} is used for zero-shot recognition.

For the bottom-up latent space learning, we conduct a comparative study on four candidate techniques (c.f. Section \ref{related_subspace}):  two unsupervised algorithms, PCA and LPP, and two supervised algorithms, LDA and SLPP\footnote{The implementation of PCA and LDA used in our experiments is based on the open source available online: \url{http://www.cad.zju.edu.cn/home/dengcai/Data/DimensionReduction.html}.}.
For fairness, 
we apply the same cross-validation procedure described in Section \ref{sub:hyper} to find out the optimal hyper-parameter values, i.e., $d_y$ for PCA, $\alpha$, $d_y$ and $k_G$ for LPP.
For LDA, however, the dimension of the latent space is intrinsically determined by the number of training classes. Hence, the dimension of its latent space is set to the number of training classes subtracted by one. Furthermore, we apply our LSM algorithm directly to visual representations without the bottom-up learning. This experiment yields a baseline that clearly exhibits the role played by each of two subsequent learning stages in our framework.

In addition, some existing ZSL methods could be enabling techniques applied to our bottom-up latent space learning\footnote{An anonymous reviewer pointed out this fact and suggested this experiment.}, e.g.,  SJE \citep{akata2015evaluation}, LatEm \citep{xian2016latent} and CCA \citep{fu2015transductive}.
Unlike the aforementioned subspace learning where no semantic representations of labels are considered, those ZSL algorithms take into account semantic representations during projection learning. For example, SJE \citep{akata2015evaluation} learns a projection matrix $W$ such that given a pair of visual and semantic representations, $\pmb{x}$ and $\pmb{y}$, similarity score $\pmb{x}^TW\pmb{y}$ is maximized if $\pmb{x}$ has a label represented by $\pmb{y}$. LatEm extends SJE to a nonlinear model with multiple piecewise linear models by learning different projection matrices  such that different instances can select the most appropriate projection matrices. CCA is an algorithm used to learn a common space from two multidimensional variables such that  the correlation between the projections of the two variables in the common space can be maximized. Furthermore, the canonical correlation problem may be converted into a distance minimization problem: $min_{W,W'} || XW - YW'||_F$ \citep{hardoon2004canonical}, where $||\cdot|||_F$ is the Frobenius norm and $W$ and $W'$ are projection matrices for source and target embedding (to the common space). In our experiments, we strictly follow the experimental setting described in the original literature and the learned projections from visual to target space are used to form the latent space. As a result, the dimensionality of the latent space is  equal to the dimensionality of semantic representations for SJE and LatEm, and 
the dimension of latent space learned by CCA is found by the same cross-validation procedure described in Section \ref{sub:hyper}.
It is worth mentioning that LatEm yields multiple projection matrices, which results in multiple ``latent" spaces. Hence, zero-shot recognition has to take into account all of such ``latent" spaces. There are two manners for the nearest-neighbor based decision-making: minimum distance and averaging distance to a label embedding in multiple ``latent" spaces. As the averaging distance always outperforms the minimum distance, we only report the results based on the averaging distance.

Our LSM algorithm described in Section \ref{sub:top-down} is always employed for the top-down embedding learning in all the aforementioned experiments regarding the bottom-up learning. We further conduct an experiment by employing the \emph{support vector regression} (SVR) \citep{smola1997support} to replace the LSM for the top-down learning. This experiment is based on SLPP used in the bottom-up stage. When SVR is used, the top-down learning is formulated as a regression task \citep{changpinyo2016predicting} and the regressor is trained based on training data where the landmarks are targets used for learning. As our LSM and the SVR work in a quite different manner for the top-down learning, it is possible to combine their results to improve the zero-shot recognition performance as well as to understand their behavior.  To this end, we further use a simple ensemble strategy to combine the two methods. Let $\pmb{b}^u_{lsm}$ and $\pmb{b}^u_{svr}$ ($u=1,\cdots, |\mathcal{C}^u|$) denote the latent embedding for unseen classes resulting from two different top-down techniques, respectively. Thus, the combined embedding of unseen classes is defined by $(\pmb{b}^u_{lsm}+\pmb{b}^u_{svr})/2$ ($u=1,\cdots, |\mathcal{C}^u|$) to be used in zero-shot recognition.

It is worth mentioning that the optimal hyper-parameter values in various candidate techniques are also achieved via the same classwise cross-validation protocol suggested by \citet{akata2016label,zhang2016zero}.

\subsection{On the Joint Use of Multiple Semantic Representations}
\label{sub:multiSR}

The joint use of multiple semantic representations can also improve the robustness in zero-shot visual recognition \citep{akata2014zero, akata2015evaluation, xian2016latent, changpinyo2016synthesized}. Our framework allows for jointly using multiple semantic representations easily. Since our recognition process described in Algorithm \ref{alg1} requires only between-class semantic distances as inputs, we use a convex combination of semantic distance matrices to exploit the information conveyed in multiple semantic spaces.

Given attributes and word vectors used in our experiments, let $\Delta^{Att}$ and $\Delta^{WV}$ denote the corresponding semantic distance matrices achieved by using attributes and word vectors, respectively. The fused distance matrix is achieved by $\Delta = \gamma \Delta^{WV} +(1-\gamma) \Delta^{Att}$,
where $\gamma$ is in the range of (0.0, 1.0) and used to trade-off the contributions of two different types of semantic representations. In our experiments, we investigate the optimal value of $\gamma$ via a grid search by setting $\gamma = 0.1, 0.2, \cdots, 0.9$ with the classwise cross-validation.

As the aforementioned strategy for the simultaneous use of two semantic representations affects both the top-down and the bottom-up learning, we have to apply the same cross-validation protocol described in Section \ref{sub:hyper} first to find the optimal values of all other hyper-parameters, $\alpha$, $d_y$, $k_G$ and $k_{ST}$, especially for the scenario that two semantic representations are jointly used. In our experiments, we exploited experimental results on a single semantic representation to achieve those optimal hyper-parameter values. As a result, we chose the set of hyper-parameter values leading to the best \emph{averaging} accuracy  regarding two semantic representations (when used individually on a visual representation) as the optimal values. Thus, this set of optimal hyper-parameter values are fixed to be used in the subsequent classwise cross-validation that decides the optimal value of $\gamma$.

\subsection{On the Comparative Study}
\label{sub:compare}

To evaluate our proposed framework thoroughly, we conduct a comparative study by comparing ours to most of state-of-the-art zero-shot visual recognition methods on four benchmark datasets described in Section \ref{dataset}. For a fair comparison, we adopt the same experimental settings and use the optimal hyper-parameter values reported in literature so that one can clearly see the results yielded by different methods under the same conditions.

Below, we briefly describe the state-of-the-art zero-shot visual recognition methods used in our comparative study.
%%%%
\begin{easylist}[itemize]
&
\textbf{Direct Attribute Prediction (DAP)}: DAP proposed by \citet{lampert2009learning} is among those earliest methods for ZSL, which is often used as a baseline in zero-shot visual recognition \citep{al2015transfer,xu2015zero,gan2016learning}. It learns a direct mapping  from visual representation to attributes of their corresponding class labels. In deployment, the attributes associated with a test instance are predicted by the learned mapping functions. Then the label of this test instance is inferred with a probabilistic model.
&
\textbf{Indirect Attribute Prediction (IAP)}: IAP \citep{lampert2009learning} is yet another baseline ZSL method \citep{al2015transfer,xu2015zero,gan2016learning}. Unlike DAP, in deployment, IAP first predicts the probability scores of all the known classes for the test instance and then apply the known class-attribute relationship in semantic space to estimate the probability scores of attributes. With the prediction of attributes, the label of this test instance is predicted in the same way as DAP.
& \textbf{Structured Joint Embedding (SJE)}: SJE \citep{akata2014zero} learns a joint embedding space by maximizing the compatibility of visual and semantic representations $\mathbf{x}^T W \mathbf{s}$. The objective used for learning $W$ in SJE is similar to that proposed for the structured SVM parameter learning \citep{tsochantaridis2005large}.
&
\textbf{Synthesized Classifiers (Syn-Classifier)}: Syn-Classifier \citep{changpinyo2016synthesized} is a recent zero-shot object recognition method that exploits the relations between known and unseen classes in the semantic space. As a result, the so-called ``phantom" classes are explored to model the relations between known and unseen classes for ZSL.
&
\textbf{Exemplar prediction (EXEM(SynC))} \citep{changpinyo2016predicting} is yet another bidirectional latent space learning method similar to ours where PCA and SVR are used to learn the latent space and to predict the exemplars for unseen classes. Once the exemplars of unseen classes are predicted, they are treated as ideal semantic representations and Syn-Classifier \citep{changpinyo2016synthesized} is used for zero-shot recognition.
&
\textbf{Latent Embedding (LatEm)}: LatEm \citep{xian2016latent} is a non-trivial extension of SJE. Instead of learning a single mapping transformation in SJE, it learns a piecewise linear compatibility function of $K$ parameter matrices $W_i~(i=1, \cdots, K)$. Given a test instance $\mathbf{x}$, it will be labelled as the class whose semantic representation maximises $ \max \limits_{1\le i \le K} \mathbf{x}^T W_i \mathbf{s}$.
&
\textbf{Hierarchical Attribute Transfer (HAT)}: HAT \citep{al2015transfer} explores the hierarchical structures underlying the set of attributes. Based on the relations of the original attributes, additional high-level attributes are exploited to enhance the knowledge transfer.
&
\textbf{Kernel-alignment Domain-Invariant Component Analysis  (KDICA)}: KDICA \citep{gan2016learning} learns a feature transformation of the visual representations to eliminate the mismatches between different classes in terms of their marginal distributions over the input. Once the transformation is learned, the representation yielded by this transformation is used for its attribute prediction.
&
\textbf{Semantic Similarity Embedding (SSE)}: SSE \citep{zhang2015zero} learns a model that decomposes the visual and semantic representations into a mixture of known classes. Thus, all the unseen classes can be represented by such ``mixture patterns". Given a test instance, its visual representation is first decomposed into the mixture of known classes, and its ``mixture pattern" is used against all the unseen classes. A label of the class with the most similar mixture pattern is assigned to this test instance.
&
\textbf{Joint Latent Similarity Embedding (JLSE)}: JLSE \citep{zhang2016zero} is one of the latest zero-shot recognition methods. It formulates zero-shot recognition as a binary prediction problem by assigning a binary label to a pair of source and target domain instances.  The visual and semantic representations are mapped to their corresponding latent spaces via dictionary learning and the joint latent similarity embedding is learnt with a probabilistic model via a joint optimization on two latent spaces so that a pair of matched source and target domain instances can be found.
&
\textbf{Unsupervised Domain Adaptation (UDA)}: UDA \citep{kodirov2015unsupervised} is proposed to tackle the domain shift problem in zero-shot recognition by regularizing the projection learning for unseen instances with the projection learned with training data in known classes. Due to using test instances in projection learning, it is a typical transductive ZSL algorithm.
&
\textbf{Transductive Multiview - Hypergraph Label Propagation (TMV-HLP)}: TMV-HLP \citep{fu2015transductive} employs multiple visual and semantic representations to learn a common space. Heterogeneous hyper-graphs are constructed for multiple views and label propagation in zero-shot object recognition. This method is proposed especially for transductive ZSL.
&
\textbf{Ridge Regression + Nearest-Neighbor (RR+NN)}: RR+NN \citep{xu2015zero} is one of latest methods proposed for zero-shot human action recognition. In \citet{xu2015zero}, a ridge regression from visual to semantic representations is learned with the training data. Then the learned regression model is first used to map a test instance from visual to semantic spaces. Then a nearest neighbour algorithm is employed to assign a class label to this test instance in the semantic space.
&
\textbf{Manifold Regression + Self-Training + Normalized Nearest-Neighbor (MR+ST+NRM)}: MR+ST+NRM \citep{xu2015zero} is one of latest methods proposed for zero-shot human action recognition. Similar to ours, the manifold of visual space is considered to learn a smooth regression model towards enhancing the generalisation to unseen classes.  The self-training (ST) and the normalized nearest neighbour (NRM) \citep{dinu2014improving} techniques are further employed towards further improving the zero-shot recognition accuracy.
\end{easylist}

\begin{table}
\centering
\caption[]{Optimal hyper-parameter values in our approach on two object recognition datasets, corresponding to different visual and semantic representations, obtained with the cross-validation protocol described in Section \ref{sub:hyper}.
\textbf{Notation}: Vis. Rep. -- Visual representation, Sem. Rep. -- Semantic representation, Att -- Attributes, WV -- Word Vectors and Comb -- The combination of attributes and word vectors.}
\label{table_d}
\begin{lrbox}{\tablebox}
\begin{tabular}{ccc|cccc}\hline
\multirow{2}{*}{\textbf{Dataset}} & \multirow{2}{*}{\textbf{Vis. Rep.}}  & \multirow{2}{*}{\textbf{Sem. Rep. }} &  \multicolumn{4}{c}{\textbf{Hyper-parameter}} \\ \cline{4-7}
                  					& & & \textbf{$\alpha$}   & \textbf{$d_y$}  & \textbf{$k_G$}  & \textbf{$k_{ST}$} \\ \hline
\multirow{6}{*}{AwA}                & \multirow{3}{*}{GoogLeNet} & WV        & 1000 & 300   & 15    & 200 \\
                                    &                            & Att       & 1000 & 50    & 5     & 180 \\
                                    &							 & Comb		 & 1000 & 50	& 5 	& 200 \\
                                    & \multirow{3}{*}{Vgg19}     & WV        & 1000 & 300   & 10    & 160 \\
                                    &                            & Att       & 1000 & 150   & 5     & 180 \\
                                    &							 & Comb	 	 & 1000 & 150 	& 5 	& 200 \\\hline
\multirow{6}{*}{CUB-200-2011}       & \multirow{3}{*}{GoogLeNet} & WV        & 0.01 & 250   & 10    & 60  \\
                                    &                            & Att       & 10   & 100   & 30    & 40  \\
                                    &							 & Comb 	 & 10	& 100 	& 30	& 60  \\
                                    & \multirow{3}{*}{Vgg19}     & WV        & 1    & 250   & 30    & 40  \\
                                    &                            & Att       & 10   & 100   & 20    & 20  \\
                                    &							 & Comb		 & 1	& 100	& 30	& 40  \\ \hline
\end{tabular}
\end{lrbox}
\scalebox{0.9}{\usebox{\tablebox}}
\end{table}

\begin{figure*}
\centering
{\includegraphics[width = 6.5in]{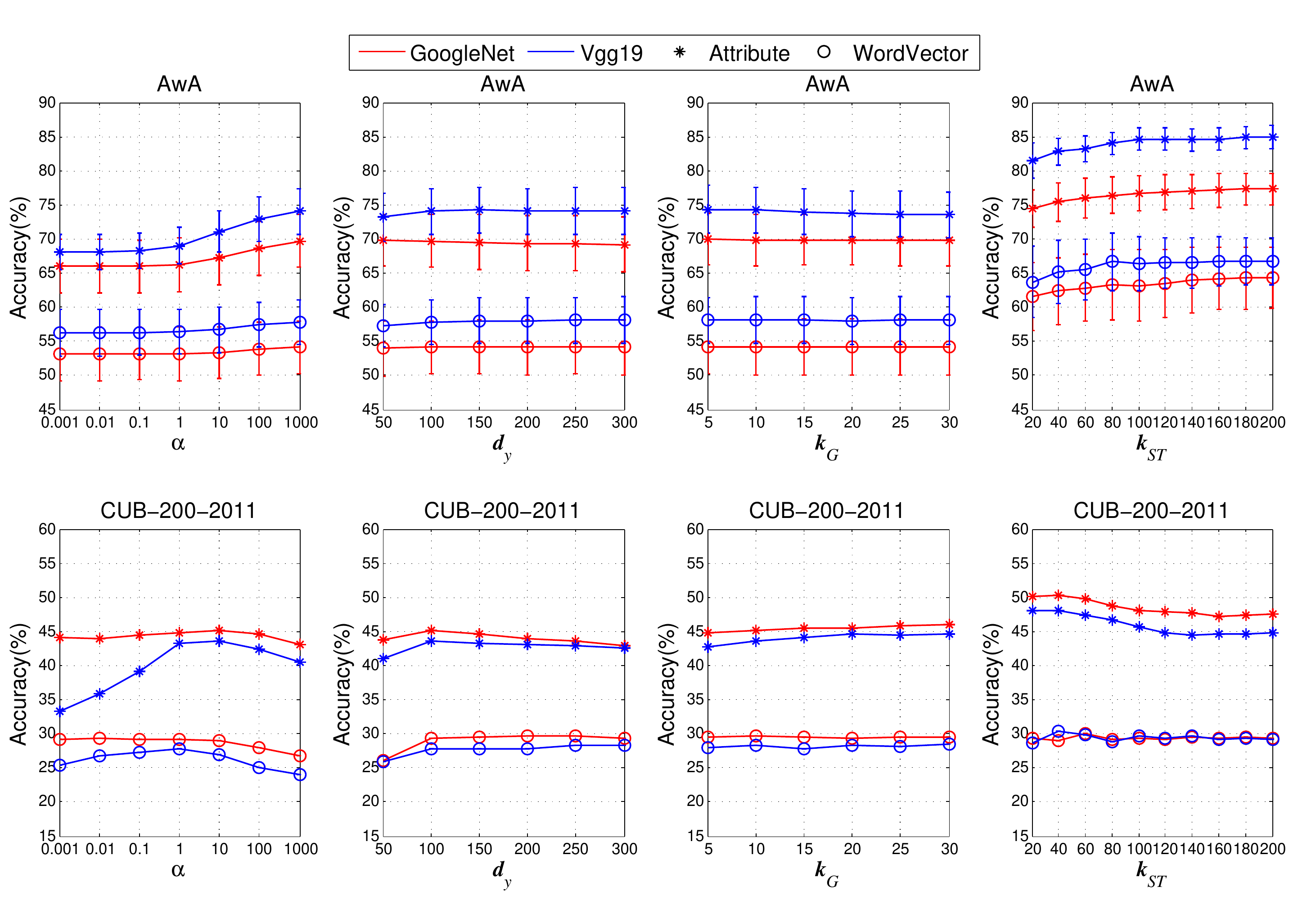}}
{\caption{The classwise cross-validation results on AwA and CUB-200-2011 used to determine the optimal hyper-parameter values.}
\label{fig_hyperparameter}}
\end{figure*}

\section{Experimental Results}
\label{results}

In this section, we report our experimental results\footnote{The source code used in our experiments as well as more experimental results not reported in this paper are available on our project website: \url{http://staff.cs.manchester.ac.uk/~kechen/BiDiLEL}.} corresponding to our settings described in Sections \ref{sub:hyper} -- \ref{sub:compare}, where the per-class accuracy is used in evaluation.

\subsection{Results on Hyper-parameters}
\label{results_hyper}

By using the cross-validation protocol described in Section \ref{sub:hyper}, we report experimental results via the mean and the standard error of per-class recognition accuracy over multiple cross-validation trials for all the datasets unless a dataset has a standard classwise split.  The initial grid search suggests that the initial optimal values of $d_y$ and $k_G$ are $100$ and $10$, respectively, regardless of different visual representations and are hence used in the hyper-parameter fine-tuning stage described in Section \ref{sub:hyper}.

\begin{table*}[th]
{\large
\centering
\caption[]{Zero-shot visual recognition performance (mean$\pm$standard error)\% of our approach resulting from the baseline without the bottom-up learning and the use of different enabling techniques in the bottom-up and the top-down learning stages.
\textbf{Notation}: Vis. Rep. -- Visual representation, Sem. Rep. -- Semantic representation, Att -- Attributes and WV -- Word Vectors.
}
\label{table_sslCpr}
\begin{lrbox}{\tablebox}
\begin{tabular}{ccc|ccccc|c|c}\hline
\multirow{2}{*}{\textbf{Dataset}} & \multirow{2}{*}{\textbf{Vis. Rep.}}  & \multirow{2}{*}{\textbf{Sem. Rep. }} &  \multicolumn{5}{c|}{\textbf{LSM}} & \textbf{SVR} & \textbf{LSM \& SVR}\\ \cline{4-10}
                               & & &\textbf{Vis. Rep.} & \textbf{PCA}   & \textbf{LPP}  & \textbf{LDA}  & \textbf{SLPP} & \textbf{SLPP} & \textbf{SLPP}\\ \hline
\multirow{4}{*}{AwA}                & \multirow{2}{*}{GoogLeNet} & WV        &  $\mathbf{57.0}$ & ${56.4}$      & ${56.2}$     & 51.1                 & ${56.1}$             & ${55.9}$ & $\mathit{57.7}$\\

                                    &                            & Att          & $\mathbf{74.2}$ & ${73.3}$      & ${72.1}$     & 72.6                 & ${72.4}$     & $\mathbf{74.1}$ & $\mathit{74.5}$\\
                                    & \multirow{2}{*}{Vgg19}      & WV           &$\mathbf{57.3}$& ${56.2}$      & ${56.4}$     & 51.0                 & ${56.7}$     & $\mathbf{57.7}$ & $\mathit{59.6}$\\
                                    &                            & Att          & $\mathbf{79.8}$ & ${78.9}$      & ${79.0}$     & 73.9                 & ${79.1}$     & 75.7 & 78.6\\ \hline
\multirow{4}{*}{CUB-200-2011}       & \multirow{2}{*}{GoogLeNet} & WV        & 29.5 & 29.3                 & 32.7                 & $\mathbf{36.7}$     & 34.5                & 30.4 & 32.7 \\
                                    &                            & Att         & 43.5 & 43.9                 & 45.9                 & 42.0                 & ${49.7}$             & $\mathbf{50.7}$ & $\mathit{52.4}$\\
                                    & \multirow{2}{*}{Vgg19}      & WV        & 29.5 & 28.9                 & 34.9                 & $\mathbf{36.7}$     & $\mathbf{37.0}$     & 33.2 & 34.9\\
                                    &                            & Att       & 42.7 & 42.8                 & 45.0                 & 42.8                 & ${47.6}$     & $\mathbf{49.7}$ & ${49.5}$\\ \hline
\multirow{6}{*}{UCF101 (81/20)}        & \multirow{2}{*}{C3D}       & WV       & $36.6\pm 1.1$ & $37.6 \pm 1.1$           & $\mathbf{38.1 \pm 1.2}$ & $31.9 \pm 0.9$             & $\mathbf{38.3 \pm 1.2}$ & ${35.1 \pm 0.8}$ & $36.5\pm 0.9$ \\
                                     &                            & Att      & $35.3\pm 1.1$ & $38.3 \pm 1.0$           & ${38.7 \pm 1.2}$ & $34.5 \pm 1.2$             & ${39.2 \pm 1.0}$ & $\mathbf{43.3 \pm 1.0}$  & $\mathit{43.7\pm 1.1}$ \\
                                    & \multirow{2}{*}{MBH}       & WV       & $21.6 \pm 0.8$ & $23.8 \pm 0.9$           & $27.3 \pm 0.9$             & $24.0 \pm 0.9$             & $\mathbf{29.9 \pm 1.1}$ & $26.6\pm 0.9$    & $27.7 \pm 0.8$ \\
                                    &                            & Att         & $21.1 \pm 0.9$ & $24.6 \pm 0.9$           & $26.5 \pm 0.8$             & $27.5 \pm 0.8$             & $\mathbf{31.4 \pm 0.8}$ & ${30.6 \pm 0.8}$  & $\mathit{32.2\pm 0.8}$ \\
                                    & \multirow{2}{*}{IDT}       & WV       & $18.4 \pm 0.5$ & $20.5 \pm 0.6$           & $28.4 \pm 0.9$             & ${31.3 \pm 1.1}$     & $\mathbf{32.6 \pm 1.1}$ & $29.4\pm 0.9$ & $31.3 \pm 1.1$\\
                                    &                            & Att         & $21.2 \pm 0.7$ & $22.9 \pm 0.8$           & $28.4 \pm 0.9$             & $\mathbf{34.5 \pm 0.9}$     & $\mathbf{34.2 \pm 0.8}$ & $33.7 \pm 0.7$ & $\mathit{35.0 \pm 0.7}$\\ \hline
\multirow{6}{*}{UCF101 (51/50)}     & \multirow{2}{*}{C3D}       & WV       & $17.8 \pm 0.4$ & $\mathbf{18.5 \pm 0.4}$ & $\mathbf{18.6 \pm 0.4}$ & $16.3 \pm 0.4$             & $\mathbf{18.9 \pm 0.4}$ & ${17.9\pm 0.5}$ & ${18.9 \pm 0.5}$ \\
                                    &                            & Att        & $18.4 \pm 0.4$ & ${20.2 \pm 0.4}$ & ${20.5 \pm 0.5}$ & $19.2 \pm 0.4$             & ${20.5 \pm 0.5}$ & $\mathbf{23.8 \pm 0.6}$  & $\mathit{24.2\pm 0.5}$ \\
                                    & \multirow{2}{*}{MBH}       & WV       & $9.7 \pm 0.3$ & $10.7 \pm 0.2$           & $12.5 \pm 0.3$             & $11.7 \pm 0.3$             & $\mathbf{14.0 \pm 0.3}$ & $12.8\pm 0.3$  & $13.5 \pm 0.3$ \\
                                    &                            & Att         & $10.0 \pm 0.3$ & $11.6 \pm 0.3$           & $12.8 \pm 0.3$             & ${14.5 \pm 0.3}$     & $\mathbf{15.2 \pm 0.3}$ &$\mathbf{15.2 \pm 0.4}$ &$\mathit{16.0 \pm 0.3}$ \\
                                    & \multirow{2}{*}{IDT}       & WV       &  $8.5 \pm 0.2$ & $9.2 \pm 0.2$           & $13.5 \pm 0.4$             & $14.4 \pm 0.4$             & $\mathbf{15.4 \pm 0.4}$ & $14.3 \pm 0.2$ & $14.9 \pm 0.3$\\
                                    &                            & Att         & $9.7 \pm 0.3$ & $10.6 \pm 0.3$           & $13.3 \pm 0.4$             & $\mathbf{17.3 \pm 0.4}$     & ${16.6 \pm 0.3}$ & ${16.5 \pm 0.4}$ & $\mathit{16.9 \pm 0.4}$\\ \hline
\multirow{3}{*}{HMDB51}             & \multirow{1}{*}{C3D}       & WV       & ${18.8 \pm 0.7}$ & ${18.5 \pm 0.7}$ & ${18.3 \pm 0.7}$ & $15.1 \pm 0.6$              & ${18.6 \pm 0.7}$ & $\mathbf{19.3 \pm 0.7}$ & $\mathit{19.5 \pm 0.6}$\\
                                    & \multirow{1}{*}{MBH}       & WV        & $10.6\pm 0.4$ & $11.7 \pm 0.4$           & $12.5 \pm 0.5$                & $12.0 \pm 0.4$             & $\mathbf{14.0 \pm 0.6}$ & $12.9\pm 0.4$ & $13.3 \pm 0.5$ \\
                                    & \multirow{1}{*}{IDT}       & WV        & $11.3 \pm 0.4$ & $10.7 \pm 0.4$           & $12.7 \pm 0.7$             & $15.4 \pm 0.5$             & $\mathbf{16.4 \pm 0.6}$ & $15.8\pm 0.6$ & $16.0 \pm 0.6$ \\
\hline
\end{tabular}
\end{lrbox}
\scalebox{0.7}{\usebox{\tablebox}}
}
\end{table*}

Fig. \ref{fig_hyperparameter} shows the detailed cross-validation results in terms of statistics (mean and standard error) obtained in the fine-tuning stage for two object recognition datasets. It is evident from Fig. \ref{fig_hyperparameter} that different values of $\alpha$ affect the recognition accuracy significantly, while $k_G$ has the least effects on performance. Based on results illustrated in Fig. \ref{fig_hyperparameter}, we choose the set of hyper-parameter values leading to the best accuracy in each case when specific visual and semantic representations work together as ``optimal" for such a case. For clarity, we explicitly list all the optimal hyper-parameter values for different scenarios on two object recognition datasets in Table \ref{table_d}. It is worth stating that the optimal hyper-parameter values for the scenario that two semantic representations are jointly used are easily achieved with the results shown in Fig. \ref{fig_hyperparameter}; i.e., for a specific visual representation, the averaging accuracy on two semantic representations can be immediately achieved at each grid point of a hyper-parameter and the optimal value can hence be found easily for this combination scenario.

As there are 30 different training/test splits \citep{xu2015semantic} for each of two human action datasets, UCF101 and HMDB51, we have 30 sets of optimal hyper-parameter values on a dataset for each of scenarios that combine specific visual and semantic representations. As we used four different visual representations and up to two semantic representations in our experiments, there are totally up to eight different scenarios. Due to the limited space, it is impossible to include all the details in this paper but we have made all the experimental results on two human action datasets available on our project website.

The optimal hyper-parameter values achieved via the aforementioned classwise cross-validation experiments are used in the comparative study reported in Section \ref{comparison}.

\subsection{Results on Enabling Techniques}
\label{results_SubLearn}

By using the settings described in Section \ref{sub:subspace}, we conduct the experiments to explore proper enabling techniques.
Table \ref{table_sslCpr} shows the zero-shot recognition performance resulting from the baseline without the bottom-up learning and the use of different enabling techniques, where a bold-font figure indicates the best performance of statistical significance in a specific setting, and a italic-font figure suggests that the performance has been improved due to the combination of  different embedding of unseen-class labels resulting from our LSM and SVR.

\begin{figure*}[th]
\centering
{\includegraphics[width = 6.5in]{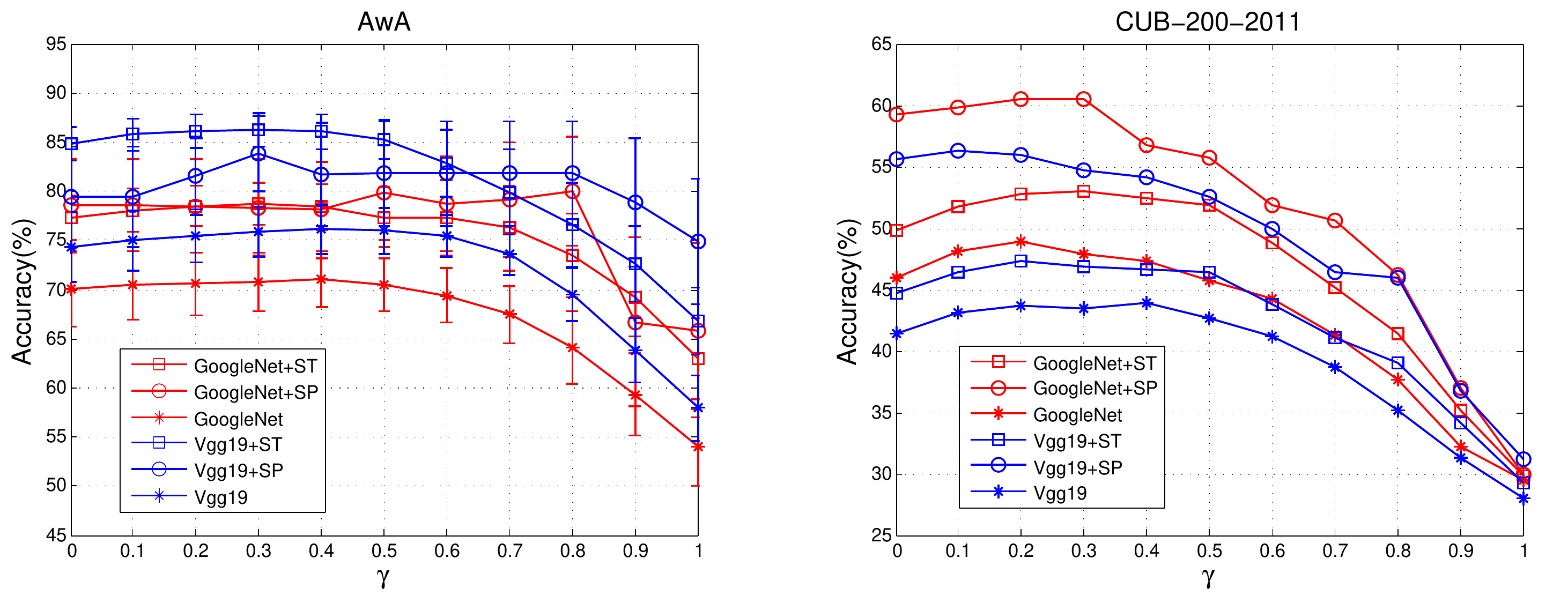}}
{\caption{The classwise cross-validation results on AwA and CUB-200-2011 when two semantic representations are jointly used.}
\label{fig_semComb}}
\end{figure*}
% Please add the following required packages to your document preamble:
% \usepackage{multirow}

Regarding those enabling techniques for the bottom-up learning, it is evident from Table \ref{table_sslCpr} that SLPP generally performs the best regardless of datasets and representations. By a closer look at Table \ref{table_sslCpr}, we observe that the performance of PCA and LPP is comparable to that of SLPP when deep representations, e.g., GoogleNet, Vgg19 and C3D, are used. This suggests that the additional use of labeling information in SLPP does not improve the generalization performance substantially. It is also evident from Table \ref{table_sslCpr} that the aggressive use of labeling information in LDA usually results in poor generalization. Such performance is attributed to the fact that, to some extent, the visual features generated by deep CNNs via supervised learning on a much larger dataset characterize the intrinsic structure of visual data and discriminative aspects of images or video streams belonging to different classes. Further supervised learning on such visual representations may lead to overfitting to training classes. It is particularly true on AwA where the deep features of visual data sufficiently capture the intrinsic ``cluster" structure; it is observed from Table \ref{table_sslCpr} that without the bottom-up learning, our LSM algorithm yields the better performance than that of itself working on four candidate subspace learning algorithms used in the bottom-up learning. This suggests that the bottom-up learning might be redundant for a dataset such as AwA. As clearly shown in Table \ref{table_sslCpr}, however, the bottom-up learning on other three datasets leads to a performance gain regardless of different visual and semantic representations used.  On the other hand, we observe that the performance of LDA is also comparable to that of SLPP when a kernel representation space is used by the joint use of multiple visual representations, e.g., IDT on UCF101. This suggests that after being mapped onto a kernel representation space, the instances in different classes are not separated well, and the use of labeling information improves the discriminative aspects in the latent space. Based on the baseline performance, we conclude that the proper bottom-up learning is required by taking into account preserving intrinsic structure underlying visual data and promoting the discriminative capability simultaneously unless a visual representation has already captured the intrinsic ``cluster" structure of a visual data set.

\begin{table}[t]
\centering
\caption[]{Results on SJE, LatEm and CCA used as the enabling techniques for the bottom-up learning while the LSM is used for the top-down learning.}

\label{table_sje}
\begin{lrbox}{\tablebox}
\begin{tabular}{cccccc}\toprule
\textbf{Dataset} & \textbf{Vis. Rep.}  & \textbf{Sem. Rep. } & \textbf{SJE}   & \textbf{LatEm}  & \textbf{CCA}  \\ \toprule
\multirow{4}{*}{AwA}                & \multirow{2}{*}{GoogLeNet} & WV       & $47.8$  	& $53.1$ & 48.9  \\
                                    &                            & Att     	& $70.0$  	& $73.2$ & 72.7 \\
                                    & \multirow{2}{*}{Vgg19} 	 & WV      	& $48.2$  	& $57.4$ & 51.9 \\
                                    &                            & Att     	& $75.7$  	& $76.5$ & 75.5 \\ 	\midrule
\multirow{4}{*}{CUB-200-2011}       & \multirow{2}{*}{GoogLeNet} & WV       & 26.8 		& 26.6 & 37.1\\
                                    &                            & Att		& 39.2 		& 34.8 & 49.7\\
                                    & \multirow{2}{*}{Vgg19} 	 & WV       & 26.7		& 25.1 & 37.9\\
                                    &                            & Att  	& 37.2 		& 36.0 & 49.2\\ \midrule
\end{tabular}
\end{lrbox}
\scalebox{0.9}{\usebox{\tablebox}}
\end{table}

Regarding the enabling top-down learning techniques, the results shown in Table \ref{table_sslCpr} reveal that LSM generally performs better than SVR, although its performance is inferior to that of SVR in some occasions for specific visual and semantic representations used on different datasets: GoogleNet+Att and Vgg19+WV on AwA, Att on CUB-200-2011 and C3D+Att on UCF101. Furthermore, an interesting phenomenon is observed from Table \ref{table_sslCpr} that the combination of LSM and SVR in unseen-class embedding always improves the performance of SVR whenever SVR outperforms LSM but the further improvement does not always happen when our LSM outperforms SVR. The experimental results exhibit the difference between the SVR, a parametric model, and our LSM, a non-parametric model in knowledge transfer.

Regarding the use of existing ZSL methods for bottom-up learning, we have only done the experiments on two object recognition benchmark datasets since results on these two datasets are only reported in the literature regarding three candidate methods, SJE, LatEm and CCA. It is evident from Table \ref{table_sje} that SLPP generally outperforms three methods on AwA
although the performance of LatEm is better than that of using specific visual and semantic representation combinations, GoogleNet+Att and Vgg19+WV.
However, CCA outperforms SLPP on CUB-200-2011 for those visual and semantic representation combinations: GoogLeNet+WV, Vgg19+WV and Vgg19+Att. This suggests that a proper enabling technique for the bottom-up learning may be dependent of a specific dataset. Fortunately, different enabling techniques can be easily and flexibly applied in our framework.

In summary, the above experimental results suggest that SLPP can preserve intrinsic structure underlying visual data and facilitate discriminating different classes in the latent space. Thus, SLPP provides a proper enabling technique for the bottom-up learning. On the other hand, our proposed LSM works effectively in comparison to SVR and is hence a proper enabling technique for the top-down learning.

% % % % % % The table of object dataset results % % % % % % % % % % % % % %

\begin{table*}[t]
{\normalsize
\centering
\caption[]{Zero-shot object recognition per-class accuracy (mean$\pm$standard deviation)\% of different approaches on AwA and CUB-200-2011 datasets.
\textbf{Notation}: Vis. Rep. -- Visual representation, Sem. Rep. -- Semantic representation, Att -- Attributes, WV -- Word Vectors, Comb -- Combination of two semantic representations. $^*$ indicates that this method uses unlabelled test instances during learning under a transductive setting. $^\dagger$ refers to the fact that the result is generated based on their specific splits publicly unavailable. $^\ddagger$ refers to the results on per-image accuracy. - refers to no result reported for this setting.}
\label{table_objectresults}
%\newsavebox{\tablebox}
\begin{lrbox}{\tablebox}
\begin{tabular}{@{}llllllll@{}}\toprule
 \multirow{2}{*}{\textbf{Method}} & \multirow{2}{*}{\textbf{Vis. Rep.}} & \multicolumn{3}{c}{\textbf{AwA}} & \multicolumn{3}{c}{\textbf{CUB-200-2011}} \\ \cmidrule(l){3-8}& & \textbf{Att} & \textbf{WV} & \textbf{Comb}  & \textbf{Att} & \textbf{WV}  & \textbf{Comb } \\ \midrule
 DAP \citep{al2015transfer} 						& GoogLeNet 	& 59.9 			& - 			& - 			& 36.7 					&- 					&-  \\

 SJE \citep{akata2015evaluation} 					& GoogLeNet 	& 66.7 			&$\mathbf{60.1}$& 73.9 			& 50.1 					& 28.4 				& $\mathbf{51.0}$ \\
 SynC \citep{changpinyo2016synthesized} 			& GoogLeNet 	& 72.9 			& - 			& 76.3 			& ${54.7}^\dagger$		& - 				& - \\
 EXEM(SynC) \citep{changpinyo2016predicting} & GoogLeNet & 77.2		 	& - 			& - 			& $\mathbf{59.8}^\dagger$ 			& - 				& - \\
 LatEm \citep{xian2016latent} 						& GoogLeNet 	& 72.5 			& 52.3 			& 76.1 			& 45.6 					& 33.1 				& 47.4 \\
 HAT \citep{al2015transfer} 						& GoogLeNet 	& 74.9 			& - 			& - 			& 51.8$^{\dagger}$ 		& - 				& - \\
 BiDiLEL(Ours) 										& GoogLeNet 	&$72.4 \pm 0.0$	&  $56.1\pm 0.0$ 			& $73.5 \pm 0.0$			& $49.7\pm 0.0$ 		& $34.5 \pm 0.0$	&$50.9 \pm 0.2$\\
 KDICA \citep{gan2016learning} 						& Vgg19 		& 73.8 			& - 			& -  			& 43.7 					& - 				& -\\
 SSE \citep{zhang2015zero} 					& Vgg19 		& $76.3 \pm 0.8$ 			& - 			& -  			& $30.4\pm 0.2$ 					& - 				& - \\
 JLSE \citep{zhang2016zero} & Vgg19 & $\mathbf{80.5\pm 0.5}^\ddagger$ & - & - & $42.1 \pm 0.6$ & - & - \\
 BiDiLEL(Ours) 										& Vgg19 		&$79.1 \pm 0.0$& $56.7 \pm 0.0$			&$\mathbf{78.8\pm 0.0}$& $47.6\pm 0.0$ 		& $\mathbf{37.0\pm 0.0}$ &$48.4\pm 0.1$\\
 \toprule
 UDA\citep{kodirov2015unsupervised}$^*$ 			& OverFeat 		& 73.2 			& - 			& 75.6 			& 39.5 					& - 				& 40.6 \\
 TMV-HLP \citep{fu2015transductive} $^*$ 			& OverFeat+Decaf& -	   			& - 			& 80.5 			& - 					& - 				& 47.9 \\
 BiDiLEL+ST (Ours)$^*$ 								& GoogLeNet 	&$86.2 \pm 0.0$& $59.5 \pm 0.0$ 			&$85.6 \pm 0.0$& $53.5\pm 0.0$& $38.0 \pm 0.0$&$56.6\pm0.0$\\
 BiDiLEL+SP (Ours)$^*$                              & GoogLeNet & $92.6 \pm 0.0$ & $\mathbf{76.0 \pm 0.0}$ & $92.5 \pm 0.0$ & $\mathbf{62.8 \pm 0.0}$ & ${37.7\pm 0.0}$ & $\mathbf{61.1\pm 0.0}$ \\
 JLSE+SP \citep{zhang2016eccv}$^*$ & Vgg19 & $92.1\pm 0.1$ & - & - & $55.3\pm 0.8$ & - & - \\
 BiDiLEL+ST(Ours)$^*$ 								& Vgg19 		&$88.5 \pm 0.0$ & $57.3 \pm 0.0$ 			& $89.7 \pm 0.0$		& $52.8\pm 0.0$			& $\mathbf{40.9 \pm 0.0}$&$53.0\pm0.0$\\
 BiDiLEL+SP (Ours)$^*$ & Vgg19 & $\mathbf{95.0 \pm 0.0}$ & $68.9\pm 0.0$ & $\mathbf{94.9\pm 0.0}$ & $59.3\pm 0.1$ & $40.6\pm 0.0$ & $57.4\pm 0.0$ \\
\bottomrule
\end{tabular}
\end{lrbox}
\scalebox{0.85}{\usebox{\tablebox}}
}
\end{table*}

\subsection{Results on the Joint Use of Multiple Semantic Representations}
\label{result_semComb}

By using the settings described in Section \ref{sub:multiSR}, we conduct experiments to seek the optimal value of $\gamma$ used in combining two semantic representations: attributes and word vectors. As there are many candidate visual representations, we adopt only those that lead to the state-of-the-art performance in our experiments. As there are no attributes available in HMDB51, our experiments are done on AwA, CUB-200-2011 and UCF101. While different values of $\gamma$ in its permissible range are used in the experiments, $\gamma=0.0$ corresponds to the situation that attributes are only used and $\gamma=1.0$ indicates that word vectors are only used.

Fig. \ref{fig_semComb} illustrates the classwise cross-validation results for different values of $\gamma$ in the joint use of two semantic representations on two object recognition datasets.  From Fig. \ref{fig_semComb}, we see the optimal hyper-parameter values for different visual representations in different settings, which are used in the comparative study reported in Section \ref{comparison}. Under the inductive setting, $\gamma=0.4$ for AwA regardless of visual representations and $\gamma=0.2,~0.4$ for CUB-200-2011 when GoogleNet and Vgg19 are used, respectively. When the self-teaching is used in the transductive setting, $\gamma=0.3$ for AwA regardless of visual representations and $\gamma=0.3,~0.2$ for CUB-200-2011 when GoogleNet and Vgg19 are used, respectively. When the structure prediction is used in the transductive setting, $\gamma=0.8,~0.3$ for AwA and $\gamma=0.3,~0.1$ for CUB-200-2011 when GoogleNet and Vgg19 are used, respectively.

Likewise, the classwise cross-validation was done on 30 training/test splits for different scenarios on each of two human action datasets, respectively, as same as described in Section \ref{results_hyper}. Consequently, those optimal $\gamma$ values on 30 splits, which are also available on our project website, are used in the comparative study reported in Section \ref{comparison}.

\subsection{Results on Comparative Study}
\label{comparison}

By using the settings described in Section \ref{sub:compare}, we conduct experiments to compare ours to a number of state-of-the-art zero-shot visual recognition methods. By using the identical experimental protocol as suggested in literature, we can directly compare the performance to that reported in literature. For our approach, we report the mean and standard deviation resulting from five random initial conditions used in the top-down learning on AwA and CUB-200-2011 as well as the mean and standard error of the mean resulting from 30 training/test splits on UCF101 and HMDB51 while the detailed experimental results can be found on our project website.
To facilitate our presentation, we group the experimental results in terms of zero-shot object and human action recognition.

\begin{table*}[t]
{\normalsize
\centering
\caption[]{Zero-shot human action recognition performance (mean$\pm$standard error)\% of different approaches on UCF101 and HMDB51 datasets.
\textbf{Notation}: Vis. Rep. -- Visual representation, Sem. Rep. -- Semantic representation, Att -- Attributes, WV -- Word Vectors, Comb -- Combination of two semantic representations. $^*$ indicates that this method uses unlabelled test instances during learning under a transductive setting.  $^\dagger$ highlights that the visual representation is encoded with bag-of-features. - refers to no result reported for this setting. }
\label{table_actionresults}
%\newsavebox{\tablebox}
\begin{lrbox}{\tablebox}
\begin{tabular}{@{}llccccccc}\toprule
  \multirow{2}{*}{\textbf{Method}} & \multirow{2}{*}{\textbf{Vis. Rep.}} & \multicolumn{3}{c}{\textbf{UCF101 (51/50)}} & \multicolumn{3}{c}{\textbf{UCF101 (81/20)}} & \textbf{HMDB51} \\ \cmidrule(l){3-9}& & \textbf{Att} & \textbf{WV} & \textbf{Comb} & \textbf{Att} & \textbf{WV} & \textbf{Comb}  & \textbf{WV}  \\ \midrule
DAP \citep{xu2015zero}         & IDT(HOG,HOF,MBH)  & $15.2\pm 0.3$ &-         & -        & -        & -        & -             & -             \\
IAP \citep{xu2015zero}         & IDT(HOG,HOF,MBH)  & $15.6\pm 0.3$ &-         & -        & -        & -        & -             & -             \\
RR+NN \citep{xu2015zero}         & IDT(HOG,HOF,MBH)     & -                &$11.7 \pm 0.2$  & -  & -             & -           & -     &$14.5\pm 0.1$    \\
DAP \citep{gan2016learning}    & C3D                & -                &-        &-         & $26.8\pm 1.1$    & -      &-       & -             \\
KDICA \citep{gan2016learning}     & C3D                 & -             &-         &-       & $31.1\pm 0.8$ & -         &-       & -             \\
BiDiLEL (Ours)                    & IDT(MBH)            & $15.2\pm 0.3$    &$14.0 \pm 0.3$  &$17.1\pm 0.3$  & $31.4\pm 0.8$    & $29.9\pm 1.1$ &$36.3\pm 1.0$   &$14.0\pm 0.6 $    \\
BiDiLEL (Ours)                     & IDT(HOG,HOF,MBH)     & $16.6\pm 0.3$    &$15.4 \pm 0.4$ & $19.5\pm 0.4$ & $34.2\pm 0.8$    & $32.6 \pm1.1$    &$39.6\pm 1.0$ &$16.4\pm 0.6$    \\
BiDiLEL (Ours)                     & C3D                 & $20.5\pm 0.5$    &$18.9 \pm 0.4$  &$24.4\pm 0.6$  & $39.2\pm 1.0$    & $38.3 \pm1.2$   &$ 47.5\pm 1.3$ &$18.6\pm 0.7$    \\
BiDiLEL (Ours)                  & C3D + IDT         & $\mathbf{22.2\pm 0.5}$ &$\mathbf{19.6 \pm 0.5}$ &$\mathbf{26.4\pm 0.6}$ & $\mathbf{43.3\pm 1.2}$ & $\mathbf{40.8 \pm1.2}$ & $\mathbf{51.1\pm 1.2}$ & $\mathbf{20.6\pm 0.8}$      \\
\toprule
UDA \citep{kodirov2015unsupervised}$^*$& IDT(MBH)$^\dagger$& $13.2\pm 0.6$ &-     &-            & $20.1\pm 1.0$ & -     & -        & -             \\
MR+ST+NRM \citep{xu2015zero}$^*$& IDT(HOG,HOF,MBH)     & -                &$18.0 \pm 0.4$ & - & -             & -            &-    &$19.1\pm 0.5$    \\
BiDiLEL+SP (Ours)$^*$            & IDT(MBH)            & $17.6\pm 0.6$    &$15.2 \pm 0.6$  &$19.1\pm 0.9$  & $41.1\pm 1.4$    & $36.6\pm 1.9$  &$44.3\pm 1.8$  &$13.5\pm 0.6$    \\
BiDiLEL+SP (Ours)$^*$             & IDT(HOG,HOF,MBH)  & $21.8\pm 0.7$ & $17.0 \pm 0.6$ &$23.3\pm 0.8$ & $48.3\pm 1.6$    & $40.3\pm 1.6$ &$51.0\pm 2.0$ &$15.9\pm 0.7$     \\
BiDiLEL+SP (Ours)$^*$             & C3D                & $28.3\pm 1.0$ &$21.4 \pm 0.8$  &$31.6\pm 1.2$  & $50.1\pm 2.0$    & $45.6\pm 2.0$ &$58.3\pm 1.8$   &$18.9\pm 1.1$    \\
BiDiLEL+SP (Ours)$^*$           & C3D + IDT   & $\mathbf{29.8\pm 1.0}$ & $\mathbf{23.0 \pm 0.9}$ &$\mathbf{35.1\pm 1.1}$ & $\mathbf{57.1\pm 1.7}$& $\mathbf{49.3\pm 2.0}$ &$\mathbf{66.9\pm 1.9}$ &$\mathbf{22.3\pm 1.1}$            \\
\bottomrule
\end{tabular}
\end{lrbox}
\scalebox{0.8}{\usebox{\tablebox}}
}
\end{table*}

\subsubsection{Results on Zero-shot Object Recognition}

Table \ref{table_objectresults} shows the performance of different approaches in zero-shot object recognition where the best performance is highlighted with bold font and the results from the inductive and the transductive settings are separated with a delimiter.

For AwA, it is evident from Table \ref{table_objectresults} that in the attribute-based inductive setting our approach based on Vgg19 visual features outperforms all other state-of-the-art approaches with a high accuracy of 79.1\%  in terms of \emph{per-class} accuracy except JLSE that reports the \emph{per-image} accuracy of 80.5\%. In its corresponding transductive setting, the use of \emph{self-training} (ST) in our approach based on  GoogLeNet and Vgg19 visual features lifts the accuracy to 86.2\% and 88.5\%, respectively, and the use of \emph{structured prediction} (SP) further improves the accuracy to 92.6\% and 95.0\%, respectively. In the word-vector based inductive setting, our approach based on Vgg19 visual features and 300-dimensional word vectors\footnote{In our experiments, we use the pre-trained 300-dimensional word vectors available online: \url{https://code.google.com/archive/p/word2vec}, where 400-dimensional word vectors are unavailable.} yields an accuracy of 56.1\%, which is lower than that of SJE but higher than that of LatEm where 400-dimensional word vectors are used in their experiments. In the transductive setting, we observe that both ST and SP lead to a higher accuracy. Especially, the use of SP dramatically improves the accuracy from 56.1\% to 76.0\% based on GoogleNet features. Our results suggest that SP is constantly superior to ST under the transductive setting. While the combination of two semantic representations significantly improves the performance of some methods, e.g., SJE, it is not a case for our approach on this dataset. It is observed that the combination of attributes and word vectors generally does not improve the performance on AwA regardless of visual representations.

For CUB-200-2011, EXEM(SynC) yields the best accuracy of 59.8\% in the attribute-based inductive setting but their classwise data split protocol is unavailable publicly. In contrast, the best performance of our approach is 49.7\% with GoogleNet features, which is better than that of DAP, LatEM, SSE, JLSE and KDICA but worse than that of SJE, HAT and SynC. The use of SP in the attribute-based transductive setting leads our approach to an accuracy of 62.8\%. In the word-vector based settings, it is evident from Table \ref{table_objectresults} that our approach outperforms all others;  37\% accuracy is achieved with Vgg19 features under the inductive setting and the use of ST and SP under the transductive setting lifts the the accuracy to 40.9\% and 40.6\%, respectively. Similar to other methods, e.g., SJE and LatEm, the joint use of two semantic representations further improves the performance of our approach on CUB-200-2011 in the inductive setting. Nevertheless, the combination of semantic representations under the transductive setting leads to limited improvement only when ST is used but does not work when SP is applied in our approach.

It is worth pointing out that the cost function used in our LSM algorithm is non-convex and the gradient-based local search only leads to a local optimum. However, our experimental results shown in Table \ref{table_objectresults} suggest that the LSM learning on two benchmark object recognition datasets is insensitive to different unseen-class embedding initialization and almost always converges to the same solution.

\subsubsection{Results on Zero-shot Human Action Recognition}

For zero-shot human action recognition, to the best of our knowledge, there are much fewer studies than zero-shot object recognition in literature. Hence, we compare ours to all the existing approaches \citep{xu2015zero, kodirov2015unsupervised, gan2016learning}. It is worth clarifying that our experiments concern only zero-shot human action recognition while the previous work \citep{xu2015zero} addresses other issues, e.g., action detection, which is not studied in our work. In addition,  \citet{xu2015zero} come up with the data augmentation technique to improve the performance. However, we notice that in their experiments, some classes from auxiliary data used for training are re-used in test, which violates the fundamental assumption of ZSL that training and test classes must be mutually excluded. Thus, we do not compare ours to theirs \citep{xu2015zero} in terms of the data augmentation. Since SP almost always outperforms ST for the post-processing, we only report the results yielded by SP under the transductive setting in Table \ref{table_actionresults}.

Table \ref{table_actionresults} shows the zero-shot recognition results of different methods on UCF101 and HMDB51. In the inductive setting, our approach yields the best performance on two different UCF101 classwise splits, 51/50 and 81/20. It is clearly seen from  Table \ref{table_actionresults} that our approach leads to the highest accuracy of 22.2\% and 19.6\% on average for the 51/50 split and the highest accuracy of 43.3\% and 40.8\% on average for the 81/20 split by using attributes and word vectors, respectively, along with appropriate visual representations. Despite the use of the same visual representations, our approach outperforms all the others regardless of semantic representations. Moreover, it is evident from Table \ref{table_actionresults} that the exactly same conclusion on the results achieved in the inductive setting can be drawn in the transductive setting, where our approach results in the highest accuracy of 29.8\% and 23.0\% on average for the 51/50 split and the highest accuracy of 57.1\% and 49.3\% on average for the 81/20 split by using attributes and word vectors, respectively, along with appropriate visual representations.
Furthermore, the results shown in Table \ref{table_actionresults} suggest that the joint use of two semantic representations always improve the performance of our approach substantially regardless of visual representations and classwise splits; for the 51/50 and the 81/20 splits, the highest accuracy  is  26.4\% and 51.1\% on average, respectively, in the inductive setting and the highest accuracy  is  35.1\% and 66.9\% on average, respectively, in the transductive setting.
For HMDB51, the behavior of our approach is identical to that on the 51/50 split of UCF101 in both inductive and transductive settings when word vectors are used. Ours yields the highest averaging accuracy of 20.6\% in the inductive setting and 22.3\% with SP along with C3D+IDT features in the transductive setting, respectively, although  our approach underperforms MR+ST+NRM when IDT(HOG,HOF,MBH) features are used.
Here, it is worth pointing out that neither of the optimal hyper-parameter search methods were described nor the detailed experimental results on each of 30 training/test splits were reported  in  \citep{xu2015zero, kodirov2015unsupervised, gan2016learning}.
In general, we summarize the main results shown in Table \ref{table_actionresults} as follows: a) the use of attributes always outperforms that of word vectors when the same visual representations are employed, which is consistent with \citep{akata2016label}; b) the deep representation C3D outperforms the state-of-the-art hand-crafted visual representations significantly in all the settings; c) the joint use of two semantic representations substantially improves the performance of our approach; and d) under the transductive setting, SP does not always improve the zero-shot recognition performance probably due to the highly complex intrinsic structure underlying visual data.

In summary, the experimental results achieved from our comparative study suggest that our proposed framework yields the favorable performance and is generally comparable to all the existing state-of-the-art zero-shot visual recognition methods described in Section \ref{sub:compare}.

\section{Concluding Remarks}
\label{conclusion}

In this paper, we have proposed a novel bidirectional latent embedding learning framework for zero-shot visual recognition. Unlike the existing ZSL approaches, our framework works in two subsequent learning stages. The bottom-up learning first creates a latent space by exploring intrinsic structures underlying visual data and the labeling information contained in training data. Thus, the means of projected training instances of the same class labels form the embedding of known class labels and are treated as landmarks. The top-down learning subsequently adopts a semi-supervised manner to embed all the unseen-class labels in the latent space with the guidance of landmarks in order to preserve the semantic relatedness between all different classes in the latent space. Thanks to the favorable properties of this latent space, the label of a test instance is easily predicted with a nearest-neighbor rule. Our thorough evaluation under comparative studies suggests that our framework works effectively and its performance is competitive with most of state-of-the-art zero-shot visual recognition approaches on four benchmark datasets.

In our ongoing research, we would further explore potential enabling techniques to improve the performance and extend our proposed framework to other kinds of ZSL problems in computer vision, e.g., multi-label zero-shot visual recognition. Despite being proposed for zero-shot visual recognition, we expect that our proposed framework also works on ZSL problems in different domains, e.g., zero-shot audio classification,  zero-shot music genre recognition and and zero-shot multimedia information retrieval.

\begin{appendices}
\renewcommand{\theequation}{A.\arabic{equation}}
\renewcommand{\thetable}{A.\arabic{table}}
\setcounter{equation}{0}
\renewcommand{\thealgorithm}{A.\arabic{algorithm}}
\setcounter{algorithm}{0}

%%%%%%%%%%%%%%%%%% Appendix A %%%%%%%%%%%%%%%%%%%%%%%%%%%%%%%%%%%%%%%%%%%%%%%%%%%%%%%%%

\section{Derivation of Gradient on the LSM Cost Function}
\label{appA}

In this appendix, we derive the gradient of $E(B^u)$ defined in Eq.(\ref{eq6}). To facilitate our presentation, we simplified our notation as follows: $d^{lu}_{ij}, d^{uu}_{ij}, \delta^{lu}_{ij}$ and
$\delta^{uu}_{ij}$ denote $d(\pmb{b}^l_i,\pmb{b}^u_j), d(\pmb{b}^u_i,\pmb{b}^u_j)$, $\delta(\pmb{s}^l_i,\pmb{s}^u_j)$ and $\delta(\pmb{s}^u_i,\pmb{s}^u_j)$, respectively, where
$d(\cdot, \cdot)$ and $\delta(\cdot, \cdot)$ are distance metrics used in the latent and semantic spaces.

Based on the simplified notation,  Eq.(\ref{eq6}) is re-written as follows:
\begin{equation}
\label{eq_gradient1}
\begin{aligned}
E(B^u)&=\frac{1}{|\mathcal{C}^l||\mathcal{C}^u|} \sum_{i=1}^{|\mathcal{C}^l|}  \frac{(d^{lu}_{ij} - \delta^{lu}_{ij})^2}{\delta^{lu}_{ij}}\\
&+\frac{2}{|\mathcal{C}^u|(|\mathcal{C}^u|-1)} \sum_{i=j+1}^{|\mathcal{C}^u|} \frac{(d^{uu}_{ij} - \delta^{uu}_{ij})^2}{\delta^{uu}_{ij}}.
\end{aligned}
\end{equation}

Let $\pmb{b}^u_{j}=(b^u_{j1},\cdots,b^u_{jd_y})$ denote the embedding of unseen class $j$ in the latent space,  where ${b}^u_{jk}$ is its $k$-th element. By applying the chain rule, we achieve
\begin{equation}
\label{eq_gradient2}
\begin{aligned}
\frac{\partial E(B^{u})}{\partial b^u_{jk}}= \frac{\partial{E(B^u)}}{\partial{d^{lu}_{ij}}} \frac{\partial{d^{lu}_{ij}}}{\partial{b^u_{jk}}} + \frac{\partial{E(B^u)}}{\partial{d^{uu}_{ij}}} \frac{\partial{d^{uu}_{ij}}}{\partial{b^u_{jk}}}.
\end{aligned}
\end{equation}
For the first term in Eq.(\ref{eq_gradient2}), we have
\begin{equation}
\label{eq_gradient3}
\frac{\partial E(B^{u})}{\partial d^{lu}_{ij}}=\frac{2}{|\mathcal{C}^l||\mathcal{C}^u|} \sum_{i=1}^{|\mathcal{C}^l|}  \frac{(d^{lu}_{ij} - \delta^{lu}_{ij})}{\delta^{lu}_{ij}},
\end{equation}
and
\begin{equation}
\label{eq_gradient4}
\frac{\partial{d^{lu}_{ij}}}{\partial{b^u_{jk}}} = \frac{-2(b^l_{ik}-b^u_{jk})}{2\sqrt{\sum_k (b^l_{ik}-b^u_{jk})^2}} = \frac{b^u_{jk}-b^l_{ik}}{d^{lu}_{ij}}.
\end{equation}
Likewise, for the second term in Eq.(\ref{eq_gradient2}), we have
\begin{equation}
\label{eq_gradient5}
\frac{\partial E(B^{u})}{\partial d^{uu}_{ij}}=\frac{4}{|\mathcal{C}^u|(|\mathcal{C}^u|-1)} \sum_{i=1}^{|\mathcal{C}^u|}  \frac{(d^{uu}_{ij} - \delta^{uu}_{ij})}{\delta^{uu}_{ij}},
\end{equation}
and
\begin{equation}
\label{eq_gradient6}
\frac{\partial{d^{uu}_{ij}}}{\partial{b^u_{jk}}} = \frac{-2(b^u_{ik}-b^u_{jk})}{2\sqrt{\sum_k (b^u_{ik}-b^u_{jk})^2}} = \frac{b^u_{jk}-b^u_{ik}}{d^{uu}_{ij}}.
\end{equation}

Inserting Eqs.(\ref{eq_gradient3})-(\ref{eq_gradient6}) into Eq.(\ref{eq_gradient2}) leads to
\begin{equation}
\label{eq_gradient5}
\begin{aligned}
\frac{\partial E(B^{u})}{\partial b^u_{jk}}&=\frac{2}{|\mathcal{C}^l||\mathcal{C}^u|} \sum_{i=1}^{|\mathcal{C}^l|}  \frac{ d^{lu}_{ij}-\delta^{lu}_{ij}} {\delta^{lu}_{ij} d^{lu}_{ij}}(b^u_{jk}-b^l_{ik})\\
&+\frac{4}{|\mathcal{C}^u|(|\mathcal{C}^u|-1)} \sum_{i=j+1}^{|\mathcal{C}^u|} \frac{ d^{uu}_{ij}-\delta^{uu}_{ij}}{\delta^{uu}_{ij} d^{uu}_{ij}} (b^u_{jk}-b^u_{ik}).
\end{aligned}
\end{equation}

Thus, we obtain the gradient of $E(B^{u})$ with respect to $B^{u}$ used in Algorithm \ref{alg1}:  $\nabla_{B^u} E(B^u)= \Big ( \frac{\partial E(B^{u})}{\partial b^u_{jk}} \Big )_{|\mathcal{C}^u| \times d_y }$.

%%%%%%%%%%%%%%%%%% Appendix B %%%%%%%%%%%%%%%%%%%%%%%%%%%%%%%%%%%%%%%%%%%%%%%%%%%%%%%%%

\section{Extension to the Joint Use of Multiple Visual Representations}
\label{appB}

In this appendix, we present the extension of our bidirectional latent embedding framework in the presence of multiple visual representations.

In general, different visual representations are often of various dimensionality. To tackle this problem, we apply the kernel-based methodology \citep{kernel-based-ML2000} by mapping the original visual space $\mathcal{X}$ to a pre-specified kernel space $\mathcal{K}$. For the visual representations $X^l$, the mapping leads to the corresponding kernel representations $K^l \in \mathbb{R}^{n_l \times n_l}$ where
$K^l_i$ is the $i$-th column of the kernel matrix $K^l$ and $K^l_{ij} = k(\pmb{x}^l_i, \pmb{x}^l_j)$. $k(\pmb{x}^l_i, \pmb{x}^l_j)$ stands for a kernel function of certain favorable properties, e.g., the linear kernel function used in our experiments is
$k(\pmb{x}^l_i, \pmb{x}^l_j)={\pmb{x}^l_i}^T\pmb{x}^l_j$.
As there is the same dimensionality in the kernel space, the latent embedding can be learned via a joint use of the kernel representations of different visual representations regardless of their various dimensionality.

Given $M$ different visual representations $X^{(1)}, X^{(2)}, ..., X^{(M)}$, we estimate their similarity matrices $W^{(1)}, W^{(2)}, ..., W^{(M)}$ with Eq.(\ref{eq2}), respectively, and generate their respective kernel matrices $ K^{(1)}, K^{(2)}, ..., K^{(M)}$ as described above. Then, we combine similarity and kernel matrices with their arithmetic averages:

\begin{equation}
\label{eqW}
\widetilde{W} = \frac{1}{M} \sum_{m=1}^{M} W^{(m)},
\end{equation}
and
\begin{equation}
\label{eqK}
\widetilde{K} = \frac{1}{M}\sum_{m=1}^{M} K^{(m)}.
\end{equation}
Here we assume different visual representations contribute equally. Otherwise, any weighted fusion schemes in \citep{yu2015kernelized} may directly replace our simple averaging-based fusion scheme from a computational perspective. However, the use of different weighted fusion algorithms may lead to considerably different performance. How to select a proper weighted fusion algorithm is non-trivial but not addressed in this paper.

By substituting $W$ and $X^l$ in Eq. (\ref{eq1}) with $\widetilde{W}$ in  Eq. (\ref{eqW}) and  $\widetilde{K}$ in Eq. (\ref{eqK}), the projection $P$ can be learned from multiple visual representations with the same bottom-up learning algorithm
(c.f. Eqs. (\ref{eq1})-(\ref{eqSol})). Applying the projection $P$ to the kernel representation of any instance leads to its embedding in the latent space. Thus, we can embed all the training instances in $X^l$ into the learned latent space by
\begin{equation}
\label{eqApplyP}
Y^l = P^T\widetilde{K}^l,
\end{equation}
where $\widetilde{K}^l$ is the combined kernel representation of training data $X^l$. For the same reason, the centralization and the $l_2$-normalization need to be applied to $Y^l$ prior to the landmark generation and the top-down learning as presented in Sections \ref{sub:bottom-up} and \ref{sub:top-down}. As the joint use of multiple visual representations merely affects learning the projection $P$, the landmark generation and the top-down learning in our proposed framework keep unchanged in this circumstance.

After the bidirectional latent embedding learning, however, zero-shot recognition described in Section  \ref{sub:recognition} has to be adapted for multiple visual representations accordingly. Given a test instance $\pmb{x}_i^u$, its label is predicted in the latent space via the following procedure. First of all, its representation in the kernel space $\mathcal{K}$  is achieved by
\begin{equation}
\widetilde{K}_i^u = \{\widetilde{k}(\pmb{x}^u_i, \pmb{x}^l_1),\widetilde{k}(\pmb{x}^u_i, \pmb{x}^l_2),...,\widetilde{k}(\pmb{x}^u_i, \pmb{x}^l_{n_l})\}^T,
\end{equation}
where $\widetilde{k}(\cdot,\cdot)$ is the combined kernel function via the arithmetic averages of $M$ kernel representations of this instance arising from its $M$ different visual representations.  Then we apply projection $P$ to map it into the learned latent space:
\begin{equation}
\pmb{y}_i^u = P^T \widetilde{K}_i^u.
\end{equation}
After $\pmb{y}_i^u$ is centralized and normalized in the same manner as done for all the training instances, its label, $l^*$, is assigned to the class label of which embedding is closest to $\pmb{y}_i^u$; i.e.,
\begin{equation}
l^* = \arg \min_l d(\pmb{y}_i^u,\pmb{b}^u_l),
\end{equation}
where $\pmb{b}^u_l$ is the latent embedding of $l$-th unseen class, and $d(\pmb{x},\pmb{y})$ is a distance metric in the latent space.

%%%%%%%%%%%%%%%%%% Appendix C %%%%%%%%%%%%%%%%%%%%%%%%%%%%%%%%%%%%%%%%%%%%%%%%%%%%%%%%%

\section{Visual Representation Complementarity Measurement and Selection}
\label{appC}

For the success in the joint use of multiple visual representations, diversity yet complementarity of multiple visual representations play a crucial role in zero-shot visual recognition.
In this appendix, we describe our approach to measuring the complementarity between different visual representations and a complementarity-based algorithm used in finding complementary visual representations to maximize the performance, which has been used in our experiments.

\subsection{The Complementarity Measurement}

The complementarity of multiple visual representations have been exploited in previous works. Although those empirical studies, e.g., the results reported by \citet{shao2016kernelized}, strongly suggest that the better performance can be obtained by combining multiple visual representations in human action classification, little has been done on a quantitative complementarity measurement. To this end, we propose an approach to measuring the complementarity of visual representations based on the diversity of local distribution in a representation space.

First of all, we define the complementarity measurement of two visual representations $X^{(1)}\in \mathbb{R}^{d_1 \times n}$ and $X^{(2)} \in \mathbb{R}^{d_2 \times n}$, where $d_1$ and $d_2$ are the dimensionality of the two visual representations, respectively, and $n$ is the number of instances. For each instance $\mathbf{x}_i, i = 1,2,...,n$, we denote its $k$ nearest neighbours ($k$NN) in space $\mathcal{X}^{(1)}$ and $\mathcal{X}^{(2)}$ by $\mathcal{N}^{(1)}_k(i)$ and $\mathcal{N}^{(2)}_k(i)$, respectively. To facilitate our presentation, we simplify our notation of $\mathcal{N}^{(m)}_k(i)$ to be $\mathcal{N}^{(m)}_i$. According to the labels of the instances in the $k$NN neighborhood, the set $\mathcal{N}^{(m)}_i$ can be divided into two disjoint subsets:
$$
\mathcal{N}^{(m)}_i = \mathcal{I}^{(m)}_i \cup \mathcal{E}^{(m)}_i, ~~m = 1,2, i = 1,2,\cdots,n
$$
where $\mathcal{I}^{(m)}_i$ and $\mathcal{E}^{(m)}_i$ are the subsets that contain nearest neighbours of the same label as that of $\mathbf{x}_i$ and of different labels, respectively. Thus, we
define the complementarity between representations $X^{(1)}$ and $X^{(2)}$ as follows:
\begin{equation}
\label{eq_comp}
c(X^{(1)}, X^{(2)}) = \frac{min(|\mathcal{I}^{(1)}|,|\mathcal{I}^{(2)}|) - |\mathcal{I}^{(1)} \cap \mathcal{I}^{(2)}|}{ |\mathcal{I}^{(1)} |+ |\mathcal{I}^{(2)} |- |\mathcal{I}^{(1)}\cap \mathcal{I}^{(2)}|},
\end{equation}
where $\mathcal{I}^{(m)} = \cup_{i=1}^n \mathcal{I}^{(m)}_i$ for  $m = 1, 2$, and  $|\cdot|$ denotes the cardinality of a set. The value of $c$ ranges from 0 to 0.5. Intuitively, the greater the value of $c$ is, the higher complementarity between two representations is.

In the presence of more than two visual representations, we have to measure the complementarity between one and the remaining representations instead of another single one as treated in Eq.(\ref{eq_comp}). Fortunately, we can extend the measurement defined in Eq.(\ref{eq_comp}) to this general scenario. Without loss of generality, we define the complementarity between representation $X^{(1)}$ and a set of representations $S = \{X^{(2)}, ..., X^{(M)}\}$ as follows:
\begin{equation}
\label{eq:c-generic}
c(X^{(1)},S)=\frac{min(|\mathcal{I}^{(1)}|,|\mathcal{I}^{2,...,M}|) - |\mathcal{I}^{(1)}\cap \mathcal{I}^{2,...,M}|}{|\mathcal{I}^{(1)}|+ |\mathcal{I}^{2,...,M}|- |\mathcal{I}^{(1)}\cap \mathcal{I}^{2,...,M}|},
\end{equation}
where $|\mathcal{I}^{2,...,M}|=|\mathcal{I}^{(2)}\cup \mathcal{I}^{(3)}...\cup \mathcal{I}^{(M)}|$. Thus, Eq. (\ref{eq:c-generic}) forms a generic complementarity measurement for multiple visual representations.

\subsection{Finding Complementary Visual Representations}

Given a set of representations $\{X^{(1)}, X^{(2)}, ..., X^{(M)}\}$, we aim to select a subset of representations $S_{selected}$ where  the complementarity between each element and another is as high as possible. Assume we already have a set $S_{selected}$ containing $m$ complementary representations, and a set $S_{candidate}$ containing $M-m$ candidate representations, we can decide which representation in $S_{candidate}$ should be selected to join $S_{selected}$ by using the complementarity measurement defined in Eq. (\ref{eq:c-generic}).  In particular, we estimate the complementarity between each candidate representation and the set of all the representations in $S_{selected}$, and the one of highest complementarity is selected. The selection procedure terminates when a pre-defined condition is satisfied. For example, a pre-defined condition may be a maximum number of representations to be allowed in $S_{selected}$ or a threshold specified by a minimal value of complementarity measurement. The complementary representation selection procedure is summarized in Algorithm \ref{alg2}.

\begin{algorithm}[htb]
\caption{Finding Complementary Representations.}
\label{alg2}
\algnewcommand\algorithmicreturn{\textbf{Return:}}
\algnewcommand\RETURN{\item[\algorithmicreturn]}
\algnewcommand\algorithmicinit{\textbf{Initialize:}}
\algnewcommand\INIT{\item[\algorithmicinit]}
\renewcommand{\algorithmicrequire}{\textbf{Input:}}
\renewcommand{\algorithmicensure}{\textbf{Output:}}
\begin{algorithmic}[1]
\REQUIRE $S_{candidate}$ and $S_{selected}=\emptyset$ .
\ENSURE $S_{selected}$.
\INIT Compute the classification performance of each representation in $S_{candidate}$, and move the one with best performance from $S_{candidate}$ to $S_{selected}$.
\WHILE {Termination condition is not satisfied}
\FOR {Each candidate representation $X^{m} \in S_{candidate}$}
\STATE Compute $c(X^{(m)}, S_{selected})$.
\ENDFOR
\STATE Select the $X^{(m)}$, with highest $c(X^{(m)}, S_{selected})$.
\STATE Move the $X^{(m)}$ from $S_{candidate}$ to $S_{selected}$.
\ENDWHILE
%\RETURN Updated $S_{selected}
\end{algorithmic}
\end{algorithm}

\subsection{Application in Zero-shot Human Action Recognition}

Here, we demonstrate the effectiveness of our proposed approach to finding complementary visual representations for zero-shot human action recognition. We apply Algorithm \ref{alg2} to candidate visual representations ranging from handcrafted to deep visual representations on UCF101 and HMDB51. For the hand-crafted candidates, we choose the state-of-the-art \emph{improved dense trajectory} (IDT) based representations. To distill the video-level representations, two different encoding methods, bag-of-features and Fisher vector, are employed to generate four different descriptors, HOG, HOF, MBHx and MBHy \citep{wang2013action}. Thus, there are a total of eight different IDT-based local representations. Besides, two global video-level representations, GIST3D \citep{solmaz2013classifying} and STLPC \citep{shao2014spatio}, are also taken into account. For deep representations, we use the C3D \citep{tran2014learning} representation. Thus, all the 11 different visual representations constitute the candidate set, $S_{candidate}$.

\begin{figure}[th]
\centering
{\includegraphics[width = 3.5in]{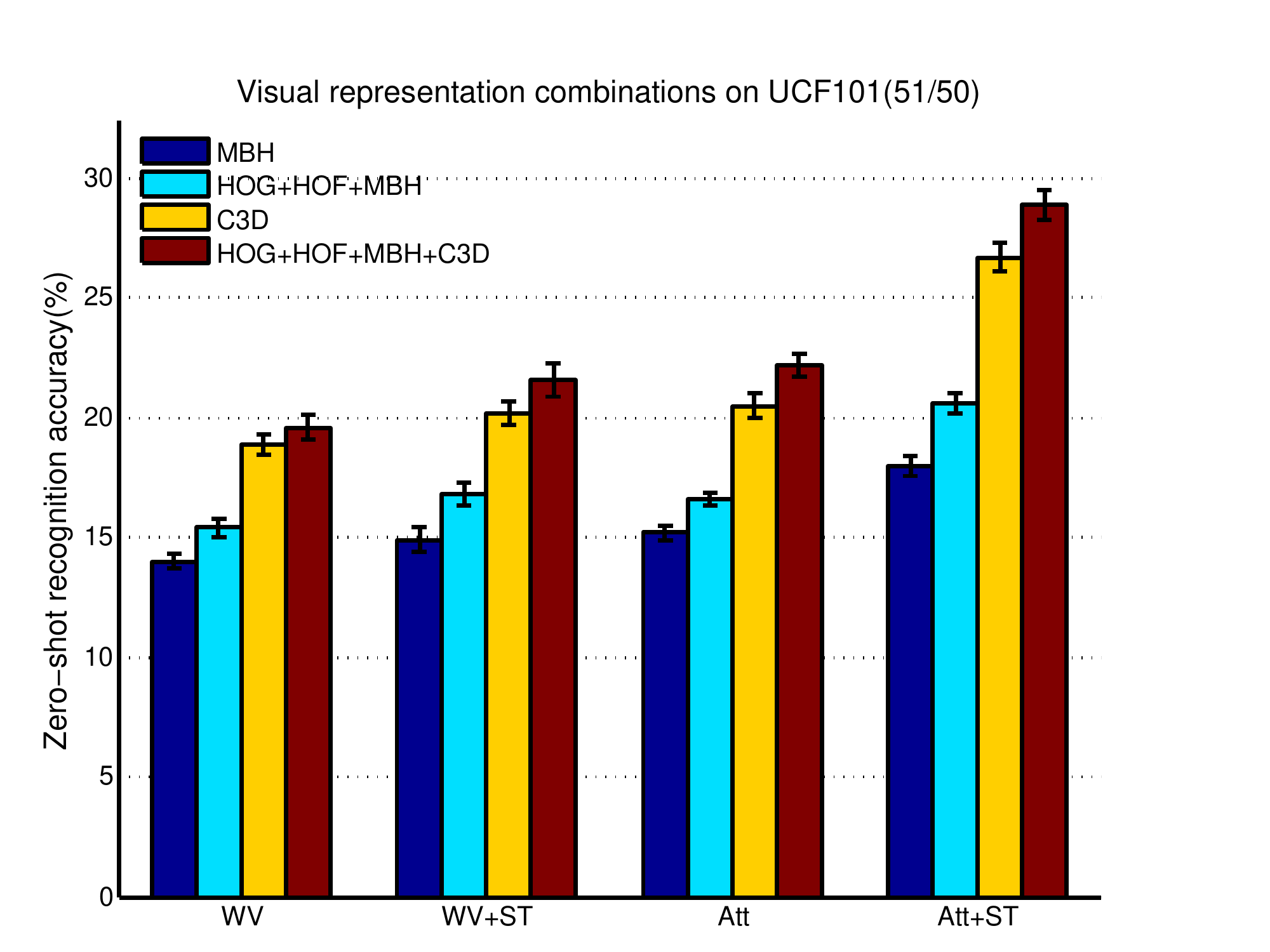}}
{\caption{Results regarding the joint use of multiple visual representations (mean and standard error) on UCF101 (51/50 split).} \label{fig_repcomb_ucf101_51}}
\end{figure}
\begin{figure}
\centering
{\includegraphics[width = 3.5in]{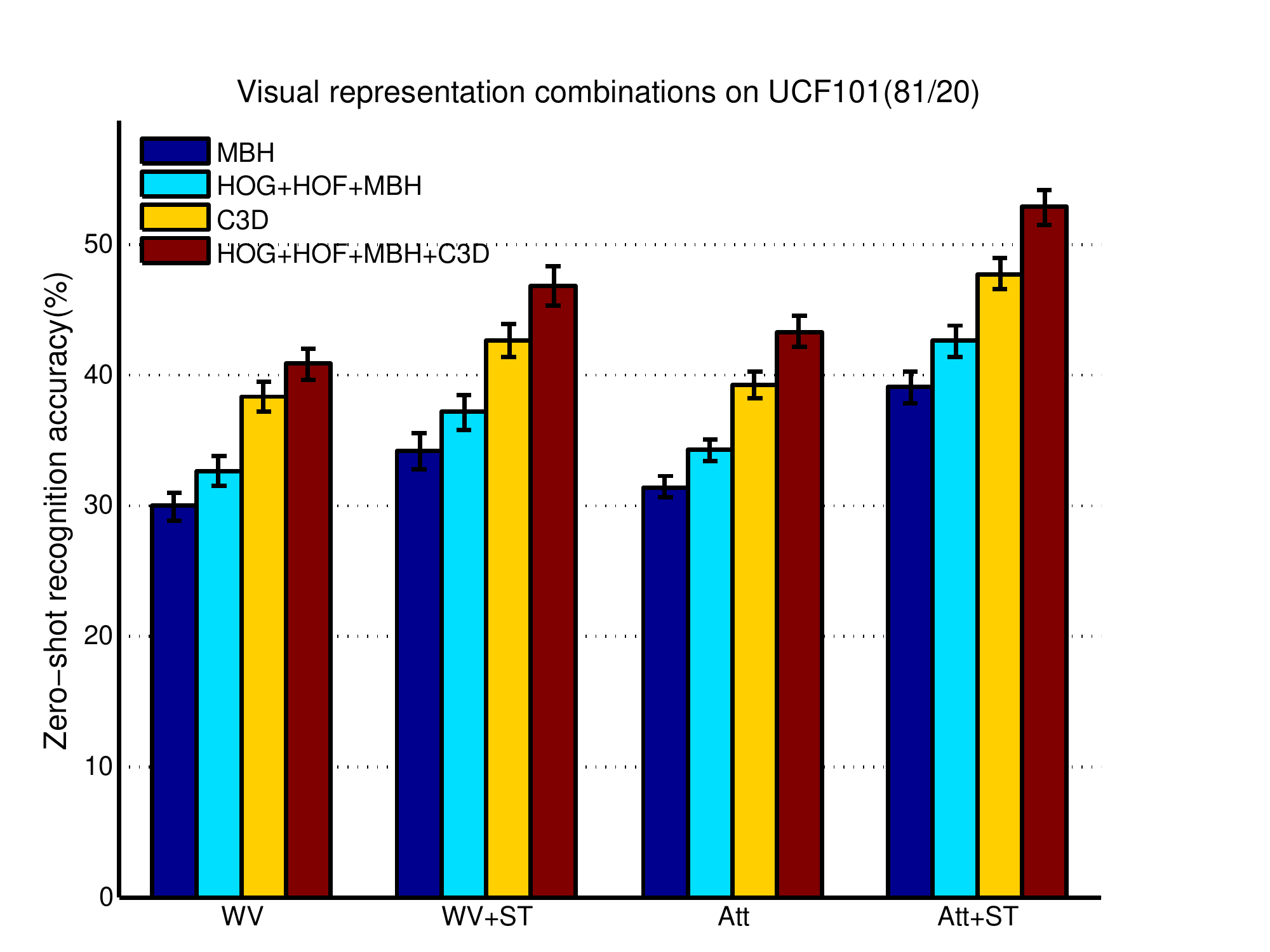}}
{\caption{Results regarding the joint use of multiple visual representations (mean and standard error) on UCF101 (81/20 split).} \label{fig_repcomb_ucf101_81}}
\end{figure}
\begin{figure}
\centering
{\includegraphics[width = 3.5in]{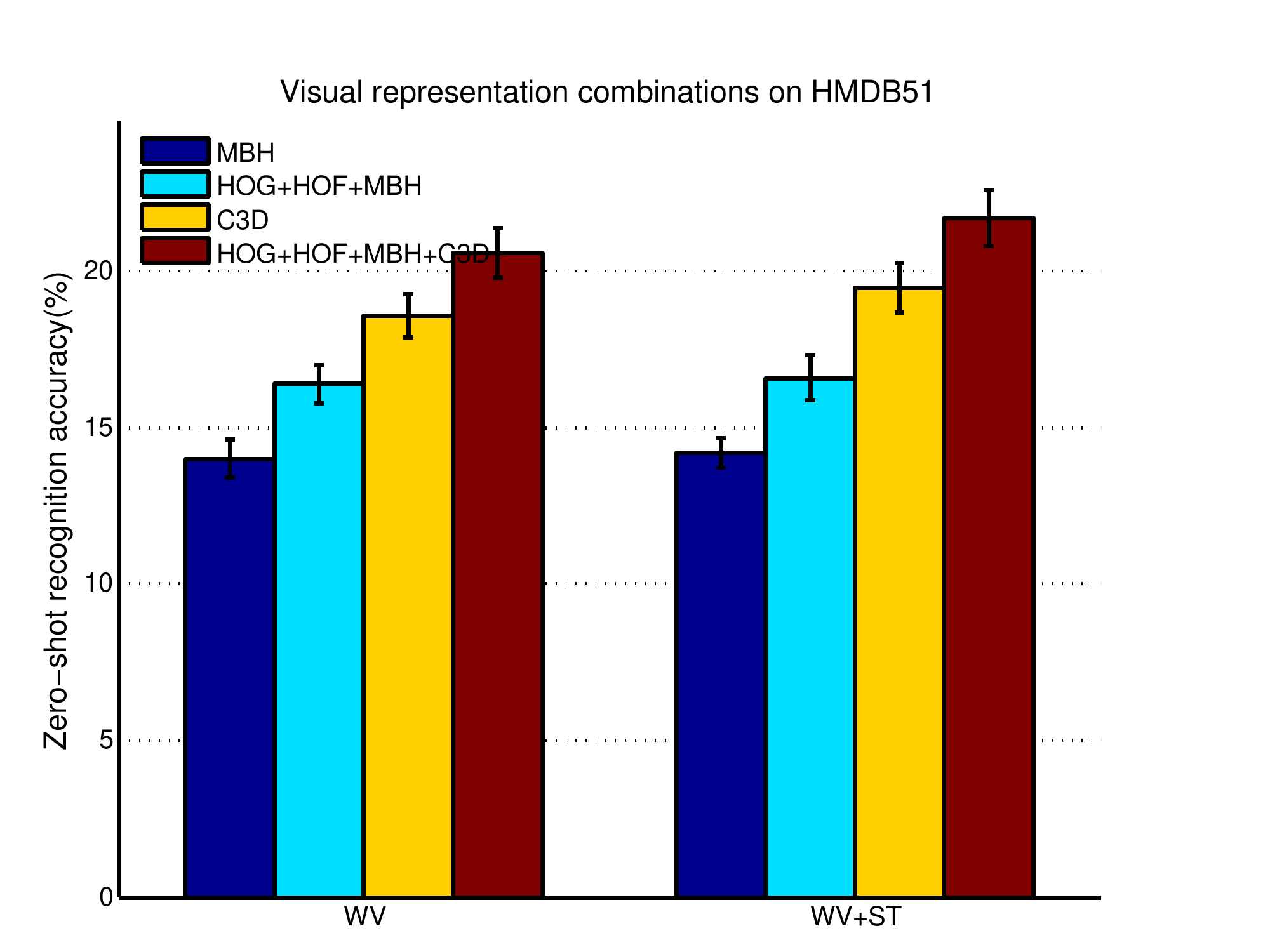}}
{\caption{Results regarding the joint use of multiple visual representations (mean and standard error) on HMDB51.} \label{fig_repcomb_hmdb51}}
\end{figure}

On UCF101 and HMDB51, we set the termination condition to be five visual representations at maximum in $S_{selected}$ in Algorithm \ref{alg2}. Applying
Algorithm \ref{alg2} to 11 candidate representations on two datasets leads to the same $S_{selected}$ consisting of C3D and four FV-based IDT representations.
To verify this measured result, we use our bidirectional latent embedding framework working on incrementally added representations with the same settings described in Section \ref{experiment}. As illustrated in Figs. \ref{fig_repcomb_ucf101_51}--\ref{fig_repcomb_hmdb51}, the performance of zero-shot human action recognition achieved in 30 trials is constantly improved as more and more selected representations are used, which suggests those selected representations are indeed complementary. In particular, the combination of the deep C3D representation and four IDT-based hand-crafted representations yields the best performance that is significantly better than that of using any single visual representations.

In conclusion, we anticipate that the technique presented in this appendix would facilitate the use of multiple visual representations in not only visual recognition but also other pattern recognition applications.

\end{appendices}

\begin{acknowledgements}
The authors would like to thank the action editor and all the anonymous reviewers for their invaluable comments that considerably improve the presentation of this manuscript. Also the authors are grateful to Ziming Zhang at Boston University for providing their source code in structured prediction and Yongqin Xian at Max Planck Institute for Informatics for providing their GoogLeNet features for AwA dataset, which have been used in our experiments.
\end{acknowledgements}

% BibTeX users please use one of
\bibliographystyle{model5-names}

\footnotesize
\bibliography{references}   % name your BibTeX data base

% Non-BibTeX users please use
%\begin{thebibliography}{}
%
% and use \bibitem to create references. Consult the Instructions
% for authors for reference list style.
%
%\bibitem{RefJ}
% Format for Journal Reference
%Author, Article title, Journal, Volume, page numbers (year)
% Format for books
%\bibitem{RefB}
%Author, Book title, page numbers. Publisher, place (year)
% etc
%\end{thebibliography}

\end{document}